\pdfoutput=1
\documentclass[11pt]{article}

\usepackage{ACL2023}

\usepackage{times}
\usepackage{latexsym}

\usepackage[T1]{fontenc}

\usepackage{amsmath}
\usepackage{amsfonts}
\usepackage{MnSymbol}

\usepackage{booktabs}
\usepackage{siunitx}
\usepackage{multirow}
\usepackage{caption}
\usepackage{subcaption}
\usepackage{pgfplots}
\pgfplotsset{compat=newest}
\usetikzlibrary{arrows.meta}
\usetikzlibrary{positioning}
\usetikzlibrary{spy}
\tikzset{>={Latex[width=3mm,length=3mm]}}
\usepackage{lscape}

\usepackage{multirow}
\usepackage{array}
\usepackage{makecell}

\newcolumntype{R}[1]{>{\raggedleft\let\newline\\\arraybackslash\hspace{0pt}}m{#1}}

\usepackage{float}
\usepackage{algorithm}
\usepackage{algpseudocode}

\algnewcommand\algorithmicforeach{\textbf{for each}}
\algdef{S}[FOR]{ForEach}[1]{\algorithmicforeach\ #1\ \algorithmicdo}

\usepackage[utf8]{inputenc}

\usepackage{microtype}

\usepackage{inconsolata}

\title{Multi-Target Cross-Lingual Summarization: a novel task and a language-neutral approach}

\author{Diogo Pernes \quad
        Gonçalo M. Correia \quad
        Afonso Mendes\\
Priberam Labs, Alameda D.\ Afonso Henriques, 41, 2º, 1000-123 Lisboa, Portugal\\
\{\href{mailto:diogo.pernes@priberam.pt}{\tt diogo.pernes},
\href{mailto:goncalo.correia@priberam.pt}{\tt goncalo.correia},
\href{mailto:amm@priberam.pt}{\tt amm}\}{\tt @priberam.pt}
}

\begin{document}
\maketitle
\begin{abstract}
Cross-lingual summarization aims to bridge language barriers by summarizing documents in different languages. However, ensuring semantic coherence across languages is an overlooked challenge and can be critical in several contexts. To fill this gap, we introduce multi-target cross-lingual summarization as the task of summarizing a document into multiple target languages while ensuring that the produced summaries are semantically similar. We propose a principled re-ranking approach to this problem and a multi-criteria evaluation protocol to assess semantic coherence across target languages, marking a first step that will hopefully stimulate further research on this problem.
\end{abstract}

\section{Introduction}

Cross-lingual summarization refers to the task of producing a summary in a different language than the original document and has the potential to break language barriers by helping people to effectively capture the essence of documents written in foreign languages \cite{wang-etal-2022-survey}. This is a very challenging task, as it combines the difficulties of monolingual summarization, such as factual inconsistencies with respect to the source document \cite{maynez-etal-2020-faithfulness}, with those of machine translation, such as translation of idiomatic expressions and cultural references \cite{fadaee-etal-2018-examining}.

The availability of large pre-trained multilingual transformers \cite{liu-etal-2020-multilingual-denoising, xue-etal-2021-mt5}, followed by the widespread development and adoption of decoder-only language models \cite{radford2018improving,touvron2023llama,jiang2023mistral,gemmateam2024gemma} has enabled a single model to perform cross-lingual summarization from multiple source languages to multiple target languages (many-to-many summarization, M2MS). Despite the increasing emphasis on this many-to-many paradigm, ensuring semantic coherence in summaries across different target languages has not been a primary focus of state-of-the-art methods, nor has it been systematically evaluated. Table~\ref{tab:summary_examples} illustrates this issue by presenting an example where a state-of-the-art M2MS system based on mT5 \cite{xue-etal-2021-mt5} produces very different summaries, with one containing unfaithful content, depending on the chosen target language. Clearly, if information is not conveyed coherently across languages, the trustworthiness of the system is compromised. Users cannot rely on the summaries to be accurate and unbiased, regardless of the language in which they consume the content. In addition, in legal or regulatory contexts, ensuring that information is presented coherently across languages can be critical. This helps meet regulatory requirements and ensures that information is transmitted coherently across language boundaries. 

\begin{table}
\begin{scriptsize}
\begin{tabular}{p{1.1cm} p{0.05cm} p{5.2cm}}
\toprule
Text (BBC)                        & en & Bitcoin uses more electricity annually than the whole of Argentina, analysis by Cambridge University suggests. "Mining" for the cryptocurrency is power-hungry, involving heavy computer calculations to verify transactions. Cambridge researchers say it consumes around 121.36 terawatt-hours (TWh) a year - and is unlikely to fall unless the value of the currency slumps. (...) \\ \midrule
\multirow{2}{*}{\makecell{\texttt{M2MS}\\ (mT5)}}             & es & Bitcoin \textcolor{red!80!black}{es la moneda más consumida del mundo}, según un estudio de la Universidad de Cambridge.                                                                                                                                                                                                                                                                                         \\
                                  & fr & Bitcoin consomme  \textcolor{red!80!black}{plus d'électricité chaque année}, selon une analyse de l'université de Cambridge.                                                                                                                                                                                                                                                                                      \\ \midrule
\multirow{2}{*}{\texttt{Mistral 7B}}          & es & Los investigadores de la Universidad de Cambridge estiman que el Bitcoin consume alrededor de 121,36 TWh de energía eléctrica al año, lo que lo convierte en un proceso energético intenso y poco probable que disminuya a menos que la valoración de la moneda baje significativamente.                                                                                               \\
                                  & fr & Des chercheurs de l'Université de Cambridge estiment que le Bitcoin consomme environ 121,36 TWh d'électricité par an, ce qui en fait un processus énergivore qui est peu probable de décroître si la valeur de la monnaie chute significativement.                                                                                                                                     \\ \midrule
\multirow{2}{*}{\makecell{\texttt{NeutralRR}\\ (ours)}} & es & Bitcoin consume más electricidad que Argentina, según un estudio de la Universidad de Cambridge.                                                                                                                                                                                                                                                                                       \\
                                  & fr & Bitcoin consomme plus d'électricité que l'Argentine, selon une analyse de l'université de Cambridge. \\ \bottomrule                                                                                                                                                                                                                                                                                  
\end{tabular}
\end{scriptsize}
\caption{An example of an en$\rightarrow$\{es, fr\} summarization task solved by three different state-of-the-art systems, including ours. Text in \textcolor{red!80!black}{red} marks information that is present in a summary for one of the languages but not in the other summary.}
\label{tab:summary_examples}
\end{table}

To fill this gap, we introduce a novel variant of cross-lingual summarization, which we call \emph{multi-target cross-lingual summarization} (MTXLS), where we specifically address the challenge of promoting semantic coherence across target languages. This framework represents an important step towards more comprehensive cross-lingual summarization techniques and evaluation. Our main contributions in this work are summarized as follows:
First, we introduce MTXLS formally as a novel task (Section~\ref{sec:mtxls}), motivated by the need of producing summaries coherently for multiple target languages. Second, we present a re-ranking-based approach to address this problem (Section~\ref{sec:methodology}), where the re-ranking phase selects a set of summaries that exhibit superior semantic coherence across target languages compared to treating each cross-lingual summarization task independently. Notably, our approach circumvents the need for a pivot language. This language-neutral strategy ensures a more robust and unbiased multilingual summarization process. Finally, we propose and conduct a multi-criteria evaluation protocol that goes beyond the simple evaluation of the similarity between generated summaries and references (Section~\ref{sec:experiments}). Specifically, we incorporate the important aspect of evaluating the coherence of the entire set of generated summaries across all target languages using quality estimation methods for machine translation. The code and data used in our experiments are publicly available.\footnote{\url{https://github.com/Priberam/MTXLSum}}

\section{Related Work}
\subsection{Cross-Lingual Summarization}
Research in cross-lingual summarization has recently gained traction, in part due to the increased availability of large datasets for this task \cite{ladhak-etal-2020-wikilingua, perez-beltrachini-lapata-2021-models, urlana-etal-2023-pmindiasum}. Among these, CrossSum \cite{bhattacharjee-etal-2023-crosssum} stands out as the most resourceful. This news dataset contains document-summary pairs for 45 different languages and more than 1,500 language directions, and it was built by automatically pairing the data from the multilingual dataset XL-Sum \cite{hasan-etal-2021-xl}, which consists of news articles from BBC.

Earlier cross-lingual summarization models operated on a per-language-pair basis \cite{zhu-etal-2019-ncls, cao-etal-2020-jointly, bai-etal-2021-cross, liang-etal-2022-variational}. However, with the emergence of large pre-trained multilingual transformers like mBART \cite{liu-etal-2020-multilingual-denoising} and mT5 \cite{xue-etal-2021-mt5}, alongside extensive cross-lingual summarization datasets covering multiple language directions, a shift to many-to-many approaches occurred \cite{bhattacharjee-etal-2023-crosssum, chen-etal-2023-revisiting, wang-etal-2023-towards-unifying}. Evaluation expanded to include large decoder-only language models, including in a zero-shot setting, with only GPT-4 showing competitive performance compared to fine-tuned mBART-50 \cite{wang-etal-2023-zero, tang-etal-2021-multilingual}. The approaches most akin to our setting in the cross-lingual summarization literature either involve first generating a summary in the source language and then using it to guide the generation of the target language summary \citep{bai-etal-2021-cross}, or employing a content plan generation step to condition the decoding of the target summary \citep{huot-etal-2024-mplan}. However, they do not explicitly enforce or evaluate semantic similarity across summaries in different target languages.

\subsection{Quality Estimation for Machine Translation}
\label{sec:qe_mt}

%In the context of machine translation (MT), quality estimation refers to the task of automatically predicting the quality of a given translation without access to gold standard output \cite{specia-etal-2013-quest, specia2018quality}. In our setting, MT quality estimation methods are particularly relevant as a tool for evaluating the semantic coherence of summaries produced by an MTXLS system across different target languages. This can be done by taking two system-generated summaries for different languages and evaluating how well one translates the other.

In machine translation (MT), quality estimation methods aim to predict translation quality without access to gold standard outputs \cite{specia-etal-2013-quest, specia2018quality}. Our focus is on using sentence-level MT quality estimation to evaluate semantic coherence in the generated summaries across target languages, by taking two system-generated summaries for different languages and evaluating how well one translates the other.

Quality estimation methods for MT can be performed at various levels: word-level, where binary labels (\texttt{OK} or \texttt{BAD}) are assigned to each machine-translated word, and sentence- or document-level, where a score is generated as an estimate of the quality of the whole translated sentence or document. Many quality estimation methods produce both word-level and sentence-level scores \cite{wang-etal-2018-alibaba, kepler-etal-2019-unbabels, kepler-etal-2019-openkiwi, lee-2020-two}. A sentence-level quality estimation method can arise from training multilingual sentence encoders like LASER \cite{artetxe-schwenk-2019-massively} or SONAR \cite{duquenne2023sonar}. These models align representations of translated sentences, allowing embedding similarity metrics in the common space to serve as quality estimation metrics for MT. BLASER \cite{chen-etal-2023-blaser}, an automatic text-free metric for evaluating speech translation, refines this idea by using a regression model trained on the concatenation of the LASER embeddings of the source text and the reference and machine-generated translations. BLASER~2.0 \cite{seamlessM4T} replaces LASER with SONAR embeddings, supports both speech and text modalities, and exists in both reference-dependent and reference-free (i.e., quality estimation) variants. Similarly, COMET~\cite{rei-etal-2020-comet} was initially introduced as a reference-dependent metric that cross-encodes the source text and the reference and machine-generated translations using an XLM-RoBERTa model \cite{conneau-etal-2020-unsupervised}. Later, a similar idea was followed to build its reference-free version, called CometKiwi \cite{rei-etal-2022-cometkiwi}.

\section{Multi-Target Cross-Lingual Summarization}
\label{sec:mtxls}

\begin{figure*}[t]
\centering
\begin{subfigure}[t]{0.3\textwidth}
  \centering
  \scalebox{.85}{\begin{tikzpicture}[auto, node distance=2cm, every loop/.style={},thick,
    main node/.style={circle,draw,font=\sffamily\bfseries,minimum size=1cm},]

    \node[main node] (xo) {$\boldsymbol{x}_o$};
    \node[main node] (pi) [right=0.5cm of xo] {$\pi$};
    \node[main node] (ypi) [below of=xo] {$\boldsymbol{y}_\pi$};
    \node[main node] (t) [right=0.5 of ypi] {$t$};
    \node[main node] (yt) [below of=ypi] {$\boldsymbol{y}_t$};

    \draw[->]
    (xo) edge (ypi)
    (pi) edge (ypi)
    (ypi) edge (yt)
    (t) edge (yt);
\end{tikzpicture}}
  \caption{}
  \label{fig:summarize_and_translate_bn}
\end{subfigure}
\begin{subfigure}[t]{0.3\textwidth}
  \centering
  \scalebox{.85}{\begin{tikzpicture}[auto, node distance=2cm, every loop/.style={},thick,
    main node/.style={circle,draw,font=\sffamily\bfseries,minimum size=1cm},]

    \node[main node] (xo) {$\boldsymbol{x}_o$};
    \node[main node] (pi) [right=0.5cm of xo] {$\pi$};
    \node[main node] (ypi) [below of=xo] {$\boldsymbol{y}_\pi$};
    \node[main node] (t) [right=0.5cm of ypi] {$t$};
    \node[main node] (yt) [below of=ypi] {$\boldsymbol{y}_t$};

    \draw[->]
    (xo) edge (ypi)
    (xo) edge[bend right] (yt)
    (pi) edge (ypi)
    (t) edge (yt);

    \draw[-]
    (ypi) edge (yt);
\end{tikzpicture}}
  \caption{}
  \label{fig:rerank_pivot_bn}
\end{subfigure}
\begin{subfigure}[t]{0.3\textwidth}
  \centering
  \scalebox{.85}{\begin{tikzpicture}[auto, node distance=2cm, every loop/.style={},thick,
    main node/.style={circle,draw,font=\sffamily\bfseries,minimum size=1cm},
    borderless node/.style={circle,font=\sffamily\bfseries,minimum size=1cm},]

    \node[main node] (xo) {$\boldsymbol{x}_o$};
    \node[main node] (yt1) [below left = 1.5cm and 1.5cm of xo] {$\boldsymbol{y}_{t_1}$};
    \node[main node] (yt2) [below left = 1.5cm and 0.1cm of xo] {$\boldsymbol{y}_{t_2}$};
    \node[borderless node] (ydots) [below right = 1.5cm and 0.1cm of xo] {$\dots$};
    \node[main node] (ytN) [below right = 1.5cm and 1.5cm of xo] {$\boldsymbol{y}_{t_N}$};

    \node[main node] (t1) [below=0.8cm of yt1] {$t_1$};
    \node[main node] (t2) [below=0.8cm of yt2] {$t_2$};
    \node[borderless node] (tdots) [below=0.8cm of ydots] {$\dots$};
    \node[main node] (tN) [below=0.8cm of ytN] {$t_N$};

    \draw[->]
    (xo) edge (yt1)
    (xo) edge (yt2)
    (xo) edge (ydots)
    (xo) edge (ytN);

    \draw[->]
    (t1) edge (yt1)
    (t2) edge (yt2)
    (tN) edge (ytN);

    \draw[-]
    (yt1) edge (yt2)
    (yt2) edge (ydots)
    (ydots) edge (ytN)
    (yt1) edge[bend right] (ydots)
    (yt1) edge[bend right] (ytN)
    (yt2) edge[bend right] (ydots)
    (yt2) edge[bend right] (ytN)
    (ydots) edge[bend right] (ytN);
\end{tikzpicture}}
  \caption{}
  \label{fig:rerank_pivot_free_bn}
\end{subfigure}
\caption{Graphical models representing summarize-and-translate (a), our method with a pivot language (b), and our language-neutral approach (c). Here, $\boldsymbol{x}_o$ denotes the document in the source language $o$, $\boldsymbol{y}_\pi$ denotes the summary in the pivot language $\pi$, and $\boldsymbol{y}_{t_i}$ denotes the summary in the target language $t_i$, $i \in \{1,2,\dots, N\}$.}
\label{fig:bayes_nets}
\end{figure*}
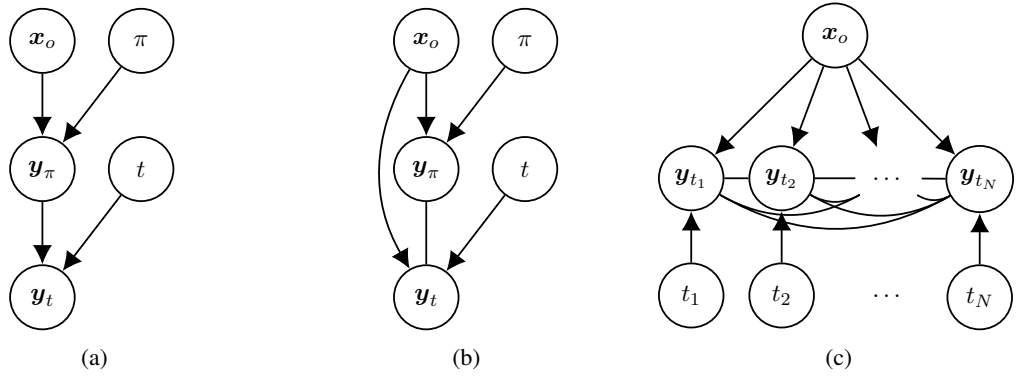

\subsection{Problem Formulation}
This section formalizes the task of MTXLS. Let $\boldsymbol{x}_o \in \mathcal{X}$ represent a document in the source language $o$, and let $\mathcal{T}=\{t_1, t_2, \dots, t_N\}$ denote a set of $N$ target languages. Without loss of generality, we assume that $o \in \mathcal{T}$. The primary goal of MTXLS is to generate a set of $N$ summaries, denoted as $\mathcal{S} = \{\boldsymbol{y}_{t_1}, \boldsymbol{y}_{t_2}, \dots, \boldsymbol{y}_{t_N}\}$, where there is a summary $\boldsymbol{y}_{t_i} \in \mathcal{Y}$ for each language in $\mathcal{T}$.

It is evident that this task can be seen as a combination of a monolingual summarization task in language $o$ and $N-1$ cross-lingual summarization tasks from $o$ to each target language $t \in \mathcal{T} \setminus \{o\}$. While these tasks could be approached independently, we impose a constraint: all $N$ summaries should convey identical information regardless of the language. This constraint ensures the alignment of information across different languages, thus promoting coherence in the resulting set of summaries.

\subsection{Summarize-and-Translate}

Consider a scenario where a summarization model is available for generating summaries from language $o$ to a pivot language $\pi$. Additionally, there are models for translating from $\pi$ to each language in $\mathcal{T}$. Common statistical approaches to these tasks involve modeling the summarization distribution $p(\boldsymbol{y}_{\pi} \mid \boldsymbol{x}_o, \pi)$ and the translation distributions $p(\boldsymbol{y}_t \mid \boldsymbol{y}_\pi, t)$ for each $t \in \mathcal{T}$.

To enforce the desired coherence constraint across target languages, a simple strategy is to assume that the target summaries are conditionally independent of the source document given the pivot summary, expressed as $(\boldsymbol{y}_t \upmodels \boldsymbol{x}_o) \mid \boldsymbol{y}_\pi , \forall t \in \mathcal{T}$ and entailed by the Bayesian network in Figure~\ref{fig:summarize_and_translate_bn}. This implies that, for each target language $t$, the information utilized to generate $\boldsymbol{y}_t$ from $\boldsymbol{x}_o$ comes solely from $\boldsymbol{y}_\pi$. Notably, since translation is a more deterministic task than summarization, this assumption serves to mitigate the potential variability of $\boldsymbol{y}_t$ across different target languages.

The previous assumption allows us to write the cross-lingual summarization distributions that use $\pi$ as the pivot language as:
\begin{align}
    p(\boldsymbol{y}_t \mid \boldsymbol{x}_o, t, \pi) &= \sum_{\boldsymbol{y}_\pi} p(\boldsymbol{y}_\pi \mid \boldsymbol{x}_o, \pi) p(\boldsymbol{y}_t \mid \boldsymbol{y}_\pi, t) \nonumber \\
    &= \mathbb{E}_{\boldsymbol{y}_\pi \mid \boldsymbol{x}_o, \pi} p(\boldsymbol{y}_t \mid \boldsymbol{y}_\pi, t),
\end{align}
for each $t \in \mathcal{T}$. Approximating this expectation with a single sample and using the source language as the pivot language yields the conventional summarize-and-translate approach to cross-lingual summarization. While this baseline ensures coherence across multiple target languages by deriving summaries from the translation of the same pivot summary, it has inherent drawbacks. In particular, it involves two successive phases of decoding: first generating the pivot summary, and then generating summaries for each target language, thus potentially suffering from error accumulation from both decoding phases. Moreover, it is likely to degrade the similarity to the reference summaries in the target languages because it is biased towards the pivot language. Thus, all resulting summaries will reflect any biases introduced during the summarization from language $o$ to language $\pi$.

\section{Methodology}
\label{sec:methodology}

\subsection{Beyond Summarize-and-Translate}
\label{sec:beyond_st}

We now relax the conditional independence assumption made previously by explicitly conditioning $\boldsymbol{y}_t$ on $\boldsymbol{x}_o$, as shown in Figure~\ref{fig:rerank_pivot_bn}. Notably, this approach does not involve decoding $\boldsymbol{y}_t$ after $\boldsymbol{y}_\pi$, but rather allows the two processes to run in parallel, and explicitly promotes semantic similarity between $\boldsymbol{y}_\pi$ and each $\boldsymbol{y}_t$, as required to satisfy our constraint. We now have:
\begin{equation}
    p(\boldsymbol{y}_t \mid \boldsymbol{x}_o, t, \pi) = \mathbb{E}_{\boldsymbol{y}_\pi \mid \boldsymbol{x}_o, \pi} p(\boldsymbol{y}_t \mid \boldsymbol{x}_o, \boldsymbol{y}_\pi, t).
\end{equation}
Let us impose that:
\begin{equation}
    p(\boldsymbol{y}_t \mid \boldsymbol{x}_o, \boldsymbol{y}_\pi, t) = \frac{1}{Z} \phi(\boldsymbol{y}_t, \boldsymbol{y}_\pi) q(\boldsymbol{y}_t \mid \boldsymbol{x}_o, t),
\end{equation}
where $Z$ is a normalizing function independent of $\boldsymbol{y}_t$, $\phi: \mathcal{Y}^2 \mapsto \mathbb{R}^+$ is a symmetric function measuring the semantic similarity between two texts in different languages and satisfies $\sum_{\boldsymbol{y}_t} \phi(\boldsymbol{y}_t, \cdot) < \infty$, and $q(\boldsymbol{y}_t \mid \boldsymbol{x}_o, t)$ is modeled by a cross-lingual summarization system from language $o$ to language $t$. This formulation explicitly addresses both of our goals: to produce a text $\boldsymbol{y}_t$ that serves as a good summary of $\boldsymbol{x}_o$ in language $t$ and has a high similarity to the pivot $\boldsymbol{y}_\pi$. Finally, we get:
\begin{align}
    p(\boldsymbol{y}_t \mid \boldsymbol{x}_o, t, \pi) &= \mathbb{E}_{\boldsymbol{y}_\pi \mid \boldsymbol{x}_o, \pi} \frac{1}{Z} \phi(\boldsymbol{y}_t, \boldsymbol{y}_\pi) q(\boldsymbol{y}_t \mid \boldsymbol{x}_o, t) \nonumber \\
    &\approx \frac{1}{Z} \phi(\boldsymbol{y}_t, \boldsymbol{y}_\pi) q(\boldsymbol{y}_t \mid \boldsymbol{x}_o, t) \nonumber \\
    &\propto \phi(\boldsymbol{y}_t, \boldsymbol{y}_\pi) q(\boldsymbol{y}_t \mid \boldsymbol{x}_o, t), \label{eq:scoring}
\end{align}
where $\boldsymbol{y}_\pi \sim p(\boldsymbol{y}_\pi \mid \boldsymbol{x}_o, t)$. This framework unveils diverse avenues for MTXLS.  One is to directly train $p(\boldsymbol{y}_t \mid \boldsymbol{x}_o, t, \pi)$ by jointly learning $\phi$ and $q$ from data, which requires cross-lingual document-summary pairs for all target languages and parallel data between the pivot and each target language. Alternatively, $\phi$ could be used as a re-scoring function at each decoding step from $q$, but this would introduce a significant computational burden.

In our work, we adopt a simpler re-ranking approach. We use $q$ to generate $k$ candidate summaries for each target language $t$, and then use $\phi$ to select the optimal candidate. Notably, this allows simultaneous generation of candidate and pivot summaries, and enhances the semantic coherence of generated summaries while maintaining similarity to the reference cross-lingual distribution used to train the summarizer, which were not possible in the summarize-and-translate approach. As shown in Section~\ref{sec:summary_sampling}, our approach has a deep connection with rejection sampling.

\subsection{A Language-Neutral Formulation}
\label{sec:language_neutral}

Despite not using translation to obtain summaries for the target languages, the approach we have described in Section~\ref{sec:beyond_st} still relies in a pivot language. However, following the same formulation, we can circumvent this issue by defining a joint distribution for the summaries in all the target languages:
\begin{equation}
p(\mathcal{S} \mid \boldsymbol{x}_o, \mathcal{T}) \propto \varphi(\mathcal{S}) \prod_{i=1}^N q(\boldsymbol{y}_{t_i} \mid \boldsymbol{x}_o, t_i),
\end{equation}
where
\begin{equation}
    \varphi(\mathcal{S}) = \frac{1}{{N \choose 2}}\sum_{i,j:\, j>i} \phi(\boldsymbol{y}_{t_i}, \boldsymbol{y}_{t_j}) \label{eq:set_similarity}
\end{equation}
measures the semantic similarity of the set of summaries $\mathcal{S}$ by averaging all the pairwise similarities between each pair of summaries in $\mathcal{S}$. This model is represented graphically in Figure~\ref{fig:rerank_pivot_free_bn}. Note that the formulation in Section~\ref{sec:beyond_st} is a particular case of this one where $\mathcal{S} = \{\boldsymbol{y}_t, \boldsymbol{y}_\pi\}$ and  $p(\mathcal{S} \mid \boldsymbol{x}_o, \mathcal{T}) = p(\boldsymbol{y}_t \mid \boldsymbol{x}_o, t, \pi) q(\boldsymbol{y}_{\pi} \mid \boldsymbol{x}_o, \pi)$.

\subsection{Summary Sampling}
\label{sec:summary_sampling}
Our primary goal is now to conceive a method that allows us to sample summaries from:
\begin{equation}
    p(\mathcal{S} \mid \boldsymbol{x}_o, \mathcal{T}) = \frac{\varphi(\mathcal{S})}{Z'} \prod_{i=1}^N q(\boldsymbol{y}_{t_i} \mid \boldsymbol{x}_o, t_i). \label{eq:set_pdf}
\end{equation}
We demonstrate we can achieve this goal through rejection sampling, which works as follows. Given a distribution $f(\boldsymbol{x})$ from which we aim to sample and a proposal distribution $g(\boldsymbol{x})$ satisfying $\sup_{\boldsymbol{x}} \frac{f(\boldsymbol{x})}{g(\boldsymbol{x})} \leq M$, we start by generating a sample $\boldsymbol{x}$ from $g$ and a sample $u$ uniformly in $[0, 1]$. Subsequently, we accept $\boldsymbol{x}$ if $\frac{f(\boldsymbol{x})}{Mg(\boldsymbol{x})} \geq u$ and reject it otherwise. 

In our context, we may use $\prod_{i=1}^N q(\boldsymbol{y}_{t_i} \mid \boldsymbol{x}_o, t)$ as the proposal distribution and assume without loss of generality that $\phi$ is bounded in $[0,1]$, so $\varphi$ is also bounded in $[0,1]$ and therefore:
\begin{equation}
    \sup_{\mathcal{S}} \frac{p(\mathcal{S} \mid \boldsymbol{x}_o, \mathcal{T})}{\prod_{i=1}^N q(\boldsymbol{y}_{t_i} \mid \boldsymbol{x}_o, t_i)} = \sup_{\mathcal{S}} \frac{\varphi(\mathcal{S})}{Z'} \leq \frac{1}{Z'}.
\end{equation}
Thus, $M = \frac{1}{Z'}$ satisfies the condition above. The rejection sampling procedure for sampling from $p(\mathcal{S} \mid \boldsymbol{x}_o, \mathcal{T})$ is then:
\begin{enumerate}
\item Sample $\mathcal{S}$ by sampling $\boldsymbol{y}_{t} \sim q(\boldsymbol{y}_{t} \mid \boldsymbol{x}_o, t)$ independently for each $t \in \mathcal{T}$.
\item Sample $u \sim U(0,1)$.
\item Accept $\mathcal{S}$ if $\varphi(\mathcal{S}) \geq u$; otherwise, reject it.
\end{enumerate}
In step 1, summaries can be sampled independently and in parallel for each target language because of the factorized form of the proposal distribution.

\subsection{A Mode-Seeking Heuristic}
\label{sec:mode_seeking}

The procedure presented in Section~\ref{sec:summary_sampling} offers a systematic means to sample sets of summaries from the distribution $p(\mathcal{S} \mid \boldsymbol{x}_o, \mathcal{T})$. However, in many practical scenarios, the objective is to obtain a single set of high-quality summaries, i.e.\ a set with high probability under this distribution. This goal motivates the approach we present here.

Let us assume we can generate $k$ candidate summaries for each target language using diverse beam search \cite{vijayakumar2018diverse} or a sampling algorithm. In this setup, there are $k^N$ different sets of summaries resulting from the different combinations of selecting a candidate from each target language. Among these sets, we wish to choose the set $\mathcal{S}^*$ that maximizes $\varphi(\mathcal{S})$, in order to achieve our goal of having a maximally semantically coherent set of summaries. Interestingly, this criterion corresponds to choosing the set $\mathcal{S}^*$ with maximum probability of being accepted in the rejection sampling procedure described in Section~\ref{sec:summary_sampling}.

However, finding $\mathcal{S}^*$ among the $k^N$ candidate sets is an instance of the generalized maximum clique problem, which is NP-hard \cite{feremans2003generalized}, and therefore we must resort to a heuristic search. For this purpose, we introduce a random permutation $\sigma$ of the target languages $\mathcal{T}$, e.g.\ $\sigma(\mathcal{T}) = (t_N, t_{N-1}, \dots, t_1)$, and define the proxy similarity function as follows:
\begin{equation}
\hat{\varphi}(\mathcal{S}; \sigma) = \frac{1}{N-1} \sum_{i=1}^{N-1} \phi(\boldsymbol{y}_{\sigma(\mathcal{T})_i}, \boldsymbol{y}_{\sigma(\mathcal{T})_{i+1}}). \label{eq:proxy_set_similarity}
\end{equation}
This proxy represents a sparsification of the clique in the graphical model shown in Figure~\ref{fig:rerank_pivot_free_bn}, where only the edges connecting adjacent target summaries according to the permutation $\sigma$ are retained. This sparsification embodies the assumption of transitivity in semantic similarity: For any three languages $a$, $b$, and $c$, if the summary $\boldsymbol{y}_a$ is similar to $\boldsymbol{y}_b$, and $\boldsymbol{y}_b$ is similar to $\boldsymbol{y}_c$, then it follows that $\boldsymbol{y}_a$ should also share a significant degree of similarity with $\boldsymbol{y}_c$. Notably, the set that maximizes $\hat{\varphi}(\mathcal{S}; \sigma)$ can be found in $O(Nk^2 \log(Nk))$ time using dynamic programming. This observation motivates Algorithm~\ref{alg:multi_target_sum}, where we consider $k$ candidate summaries per target language and $m \ll N!$ random permutations of the target languages.  Then, for each permutation, we find the candidate set $\hat{\mathcal{S}}^*_i$ that maximizes $\hat{\varphi}(\mathcal{S}; \sigma_i)$ using dynamic programming. Finally, we choose the set among $\hat{\mathcal{S}}^*_1, \hat{\mathcal{S}}^*_2, \dots, \hat{\mathcal{S}}^*_m$ that has the highest score according to $\varphi$.

\begin{algorithm}[t]
\begin{footnotesize}
\caption{Language-neutral multi-target cross-lingual summarization}\label{alg:multi_target_sum}
\begin{algorithmic}
\Require Input document ($\boldsymbol{x}_o$); Set of target languages ($\mathcal{T}$, with size $N$); Number of candidates per language $(k)$; Number of random permutations $(m)$.
\ForEach{$t \in \mathcal{T}$}  \Comment{Generate candidates}
    \For{$i \gets 1$ to $k$}
        \State Sample $\boldsymbol{y}_t^{(i)} \sim q(\boldsymbol{y}_t \mid \boldsymbol{x}_o, t)$.
    \EndFor
\EndFor
\For{$i \gets 1$ to $m$} \Comment{Find set with high similarity}
    \State Build a weighted directed graph $\mathcal{G} = (\mathcal{V}, \mathcal{E})$, where $\mathcal{V}$ has $Nk+2$ nodes, one for each candidate summary plus a \texttt{source} and a \texttt{sink} node, and $\mathcal{E} \gets \emptyset$.
    \State Sample a random permutation $\sigma(\mathcal{T}) = (t'_1, t'_2, \dots, t'_N)$.
    \State $\mathcal{E} \gets \mathcal{E} \cup \{(\texttt{source} \to \boldsymbol{y}_{t'_1}^{(i)}, 0)\}_{i=1}^k$
    \State $\mathcal{E} \gets \mathcal{E} \cup \{(\boldsymbol{y}_{t'_N}^{(i)} \to \texttt{sink}, 0)\}_{i=1}^k$
    \For{$l \gets 1$ to $N-1$}
        \State $\mathcal{E} \gets \mathcal{E} \cup \{(\boldsymbol{y}_{{t'}_l}^{(i)} \to \boldsymbol{y}_{{t'}_{l+1}}^{(j)}, 1-\phi(\boldsymbol{y}_{{t'}_l}^{(i)}, \boldsymbol{y}_{{t'}_{l+1}}^{(j)}))\}_{i,j=1}^k$
    \EndFor
    \State $\hat{\mathcal{S}}^*_i \gets \mathrm{shortest\,path}(\mathcal{G}, \texttt{source}, \texttt{sink})$
\EndFor
\State \Return $\hat{\mathcal{S}}^* \gets \mathrm{arg}\max_{\mathcal{S} \in \{\hat{\mathcal{S}}^*_1, \dots, \hat{\mathcal{S}}^*_m\}} \varphi(\mathcal{S})$ \Comment{eq. (\ref{eq:set_similarity})}
\end{algorithmic}
\end{footnotesize}
\end{algorithm}

\subsection{Choice of $\phi$}
\label{sec:choice_of_phi}
So far, we have presented our methodology in a formal manner, but have not yet provided specifics on implementing a function $\phi$ capable of measuring the semantic similarity between two summaries in different languages. In practice, any quality estimation model for MT (Section~\ref{sec:qe_mt}) could be used. In our experiments, we leverage the cosine similarity of SONAR embeddings \cite{duquenne2023sonar} as the similarity metric, reserving BLASER~2.0 \cite{chen-etal-2023-blaser} and CometKiwi \cite{rei-etal-2022-cometkiwi} for evaluation. Our selection of the cosine similarity of SONAR embeddings is motivated by its symmetry, unlike the remaining options, and the fact that the SONAR encoder is relatively lightweight. Specifically, we define the similarity function as:
\begin{equation}
    \phi(\boldsymbol{y}_a, \boldsymbol{y}_b) = \frac{1 + \boldsymbol{s}_a^\top \boldsymbol{s}_b}{2},
\end{equation}
where $\boldsymbol{s}_a$ and $\boldsymbol{s}_b$ represent the $L_2$-normalized SONAR embeddings of summaries $\boldsymbol{y}_a$ and $\boldsymbol{y}_b$.

\section{Experiments}
\label{sec:experiments}

\setlength{\tabcolsep}{3pt}
\begin{table*}[h!]
\small
\centering
\begin{tabular}{ll rrr rrr
 rrr  rrr rr} \toprule
\multirow{2}{*}{Source} & \multirow{2}{*}{Method} & \multicolumn{3}{c}{ROUGE-2 (R)}                                                                                                                         & \multicolumn{3}{c}{BLASER 2.0 (R)} & \multicolumn{3}{c}{CometKiwi (C)} & \multicolumn{3}{c}{BLASER 2.0 (C)} & \multicolumn{1}{c}{\multirow{2}{*}{T}} & \multicolumn{1}{c}{\multirow{2}{*}{\#P}} \\
                        &                         & \multicolumn{1}{c}{en} & \multicolumn{1}{c}{zh} & \multicolumn{1}{c}{rest} & \multicolumn{1}{c}{en} & \multicolumn{1}{c}{zh} & \multicolumn{1}{c}{rest} & \multicolumn{1}{c}{en} & \multicolumn{1}{c}{zh} & \multicolumn{1}{c}{rest} & \multicolumn{1}{c}{en} & \multicolumn{1}{c}{zh} & \multicolumn{1}{c}{rest} &  & \\ \midrule
\multirow{5}{*}{en} & \texttt{M2MS} & \textbf{17.88} & \textbf{18.63} & \textbf{13.20} & \hphantom{ }\underline{3.52} & \hphantom{ }3.04 & \hphantom{ }3.25 & 59.28 & 61.00 & 60.31 & \hphantom{ }3.48 & \hphantom{ }3.26 & \hphantom{ }3.61 & 0.52 & 582 \\
                    & \texttt{S\&T} & \textbf{17.88} & 7.51 & 11.77 & \underline{3.52} & 2.73 & 3.26 & \textbf{85.00} & \textbf{79.24} & \textbf{86.40} & \textbf{4.67} & \textbf{3.87} & \textbf{4.79} & 0.53 & 1,953 \\
                    & \texttt{Mistral 7B} & 6.52 & 3.18 & 4.57 & 2.45 & 2.13 & 2.31 & \underline{69.77} & \underline{65.95} & \underline{71.09} & 3.09 & 3.24 & 3.16 & 4.64 & 7,241 \\
                    & \texttt{PivotRR} \scriptsize{(ours)} & \textbf{17.88} & 17.54 & \underline{13.12} & \underline{3.52} & \textbf{3.09} & 3.28 & 63.72 & 64.42 & 63.35 & 3.71 & 3.45 & 3.81 & 0.96 & 1,348 \\
                    & \texttt{NeutralRR} \scriptsize{(ours)} & \underline{17.59} & \underline{17.87} & 12.90 & \textbf{3.53} & \underline{3.08} & \textbf{3.29} & 64.34 & 65.43 & 64.76 & \underline{3.76} & \underline{3.49} & \underline{3.89} & 0.99 & 1,348 \\ \midrule
\multirow{5}{*}{zh} & \texttt{M2MS} & 17.95 & \textbf{24.13} & 16.32 & 3.58 & \underline{3.14} & 3.31 & 61.95 & 60.23 & 60.56 & 3.40 & 3.20 & 3.39 & 0.40 & 582 \\
                    & \texttt{S\&T} & 13.51 & \textbf{24.13} & 12.11 & 3.48 & \underline{3.14} & 3.25 & \textbf{83.61} & \textbf{82.50} & \textbf{82.09} & \textbf{4.26} & \textbf{4.10} & \textbf{4.29} & 0.52 & 1,953 \\
                    & \texttt{Mistral 7B} & 4.58 & 3.93 & 3.68 & 2.47 & 2.02 & 2.39 & \underline{67.28} & \underline{66.40} & \underline{66.98} & 3.19 & 2.98 & 3.15 & 11.48 & 7,241 \\
                    & \texttt{PivotRR} \scriptsize{(ours)} & \underline{18.32} & \textbf{24.13} & \underline{16.36} & \underline{3.60} & \underline{3.14} & \textbf{3.36} & 64.73 & 62.99 & 61.90 & 3.54 & 3.37 & 3.54 & 0.89 & 1,348 \\
                    & \texttt{NeutralRR} \scriptsize{(ours)} & \textbf{18.34} & \underline{23.72} & \textbf{16.37} & \textbf{3.61} & \textbf{3.18} & \underline{3.35} & 66.94 & 63.74 & 63.23 & \underline{3.63} & \underline{3.43} & \underline{3.62} & 0.90 & 1,348 \\ \midrule
\multirow{4}{*}{rest} & \texttt{M2MS} & \textbf{16.73} & \textbf{23.83} & \textbf{13.83} & 3.48 & \underline{3.07} & 3.23 & 60.50 & 60.33 & 61.13 & 3.55 & 3.15 & 3.54 & 0.56 & 582 \\
     & \texttt{S\&T} & 11.88 & 7.63 & 11.41 & 3.38 & 2.72 & 3.20 & \textbf{85.63} & \textbf{80.38} & \textbf{85.67} & \textbf{4.71} & \textbf{3.88} & \textbf{4.75} & 0.59 & 1,953\\
     & \texttt{PivotRR} \scriptsize{(ours)} & 16.32 & \underline{23.56} & 13.66 & \underline{3.50} & \textbf{3.12} & \underline{3.25} & 63.63 & 62.04 & 63.28 & 3.72 & 3.33 & 3.73 & 0.98 & 1,348 \\
     & \texttt{NeutralRR} \scriptsize{(ours)} & \underline{16.48} & 23.01 & \underline{13.75} & \textbf{3.51} & \textbf{3.12} & \textbf{3.27} & \underline{65.37} & \underline{63.30} & \underline{64.62} & \underline{3.83} & \underline{3.39} & \underline{3.82} & 1.02 & 1.348\\ \bottomrule
\end{tabular}
\caption{Results of evaluated methods in CrossSum for multi-target cross-lingual summarization using different languages as the source language. The language in each column is the target, with ``rest" indicating the average for the remaining target languages. Metrics with (R) evaluate similarity to reference summaries, while those with (C) evaluate semantic coherence across languages. ROUGE-2 and CometKiwi range from 0 to 100, while BLASER~2.0 ranges from 1 to 5 (higher values are better). Best results are bold, second best results are underlined. Columns T and \#P indicate the average computation time per generated summary in seconds and the number of model parameters in millions, respectively.}
\label{tab:main_results}
\end{table*}
\setlength{\tabcolsep}{6pt}

\subsection{Dataset}
\label{sec:dataset}
We use a subset of the CrossSum dataset, including data in the following languages: Arabic, Chinese (simplified), English, French, Portuguese, Russian, and Spanish. CrossSum pairs documents in one language with summaries from documents in another language, using automatic similarity metrics. However, mispairings are frequent due to this automated process. Additionally, the dataset is designed for single-target cross-lingual summarization and does not perfectly fit our multi-target setting. To adapt the dataset to our needs, we restructured the dataset into clusters. This process is explained in Appendix~\ref{app:dataset_analysis}. Each resulting cluster consists of up to seven multilingual document-summary pairs, with one such pair for each language. This allows us to select any document within the cluster as a source for summarization, with all summaries within the cluster serving as references for each of the languages. Statistics about the clustered data and an analysis of the semantic coherence of the dataset summaries are also provided in Appendix~\ref{app:dataset_analysis}.

\subsection{Methods}
\label{sec:methods}
Our pivot-free re-ranking method (\texttt{NeutralRR}) proposed in Algorithm~\ref{alg:multi_target_sum} was tested using $k=$ 8 candidates per target language for re-ranking and $m=$ 6 language permutations, unless otherwise specified. We study the effects of varying $k$ and $m$ in Section~\ref{sec:exp_num_candidates} and Appendix~\ref{app:exp_heuristic_search}, respectively. We compare our method with four other approaches, namely: a many-to-many summarizer with beam search decoding (\texttt{M2MS}) with a beam size of 8; the summarize-and-translate approach (\texttt{S\&T}), where summaries are obtained in the source language and then translated to each of the target languages using beam search with a beam size of 8 in both decoding steps; a \texttt{Mistral 7B} \cite{jiang2023mistral} large language model (LLM) used in a zero-shot setting and instructed to write summaries with identical information for all the target languages (see Appendix~\ref{app:llm_prompt}); our pivot-dependent re-ranking approach (\texttt{PivotRR}) as described in Section~\ref{sec:beyond_st}, where we use the source language as the pivot.

All summaries except those of \texttt{Mistral 7B} were decoded from the same mT5 base model \cite{xue-etal-2021-mt5} fine-tuned in CrossSum. In the \texttt{S\&T} approach, translations were performed using the NLLB 1.3B model \cite{costa2022no}. \texttt{NeutralRR} and \texttt{PivotRR} used beam search multinomial sampling using with 5 beams and a temperature of 1.0 for candidate generation.\footnote{\url{https://huggingface.co/docs/transformers/generation_strategies\#beam-search-multinomial-sampling}} The pivot summary in \texttt{PivotRR} was decoded using beam search with 8 beams. For \texttt{Mistral 7B}, we used multinomial sampling with a temperature of 0.1. Further implementation details are provided in Appendix~\ref{app:imp_details}.

\subsection{Evaluation Metrics}
\label{sec:metrics}

Throughout this work, we emphasize the importance of evaluating MTXLS not only by comparing the generated summaries for each target language with their respective references, but also by evaluating the semantic coherence across different target languages. To evaluate the former, we present the ROUGE-2 F1 scores \cite{lin-2004-rouge} for each generated summary against its corresponding reference in the same target language. In addition, we calculate the BLASER~2.0 score \cite{seamlessM4T} by treating the generated summary as the translation and the reference summary for the source language as the source text. This evaluation metric is justified due to mismatched articles in CrossSum, as explained in Section~\ref{sec:dataset}, which reduces the reliability of reference summaries in languages other than the source.

To assess semantic coherence across various target languages, we evaluate how well each generated summary translates the generated summaries for the remaining target languages. For this purpose, we use two quality estimation models for MT, namely CometKiwi \cite{rei-etal-2022-cometkiwi} and BLASER~2.0. Here, for each target language, we use the generated summary as the translation and the summaries generated for all the other target languages as the source texts and then report the average across those languages.

\subsection{Main Results}
\label{sec:main_results}

\begin{figure*}[t]
\centering
\includegraphics[width=1.0\textwidth]{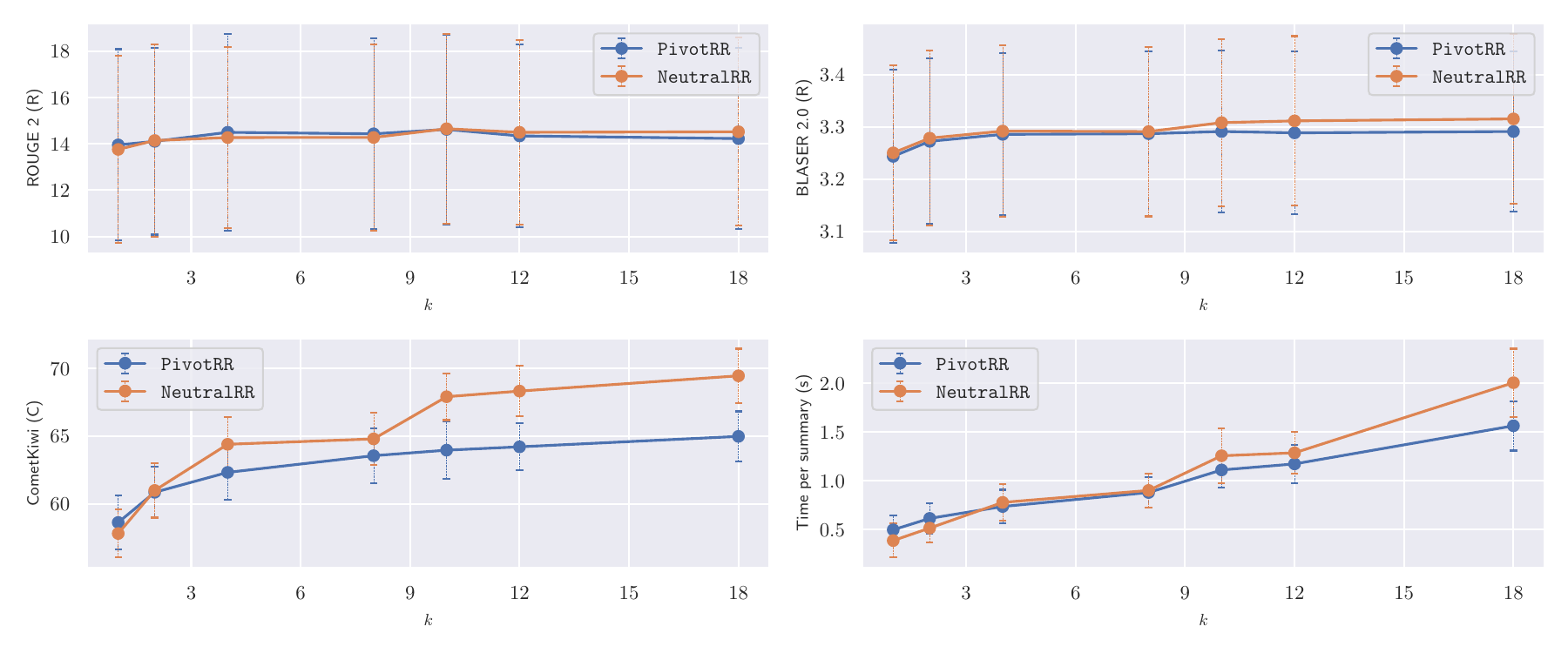}
    \caption{Results of \texttt{PivotRR} and \texttt{NeutralRR} as a function of the number of candidates per target language for re-ranking ($k$ in Algorithm~\ref{alg:multi_target_sum}). Error bars indicate the standard deviations across the target languages.}
    \label{fig:results_multicand}
\end{figure*}

In this section, we present results on MTXLS considering all the seven languages mentioned in Section~\ref{sec:dataset} as targets. To perform this task, we took each of the seven languages as the source in turn and discarded the clusters that lacked a document in the source language. Then, we iterated through the remaining clusters taking the document in the source language as the input for summarization and we generated summaries for all the languages in the cluster, including the source language, using each of the methods mentioned in Section~\ref{sec:methods}.

The results are in Table~\ref{tab:main_results} and are presented per language pair. Due to space limitations, we present detailed results only for English (en) and Chinese (zh), and show the averages for the remaining source and target languages (rest). An extended version of this table, including detailed results for more languages, confidence intervals, and the accuracy of each approach on following the target language is shown in Appendix~\ref{app:main_results_extended}. When the source and target languages are the same, \texttt{S\&T} and \texttt{PivotRR} reduce to \texttt{M2MS} because we use the source language as the pivot. Consequently, the results of these three methods for ROUGE-2 and BLASER~2.0 (R) coincide for en$\rightarrow$en and zh$\rightarrow$zh.

We begin by discussing the results of \texttt{Mistral 7B}, as these deserve special attention. Interestingly, the model always performs worst in terms of similarity to the reference summaries (ROUGE-2 and BLASER~2.0 (R)), even though it was instructed that the articles were obtained from the BBC and that the summaries should follow the BBC style (see Appendix~\ref{app:llm_prompt}). Regarding the coherence across target languages, we observe that the model has a very decent performance, as illustrated in Table~\ref{tab:summary_examples}, ranking second in CometKiwi scores, only surpassed by \texttt{S\&T}. However, the model often failed to produce the output in the requested format, in which case we had to repeat the request, or did not produce text in the specified target language (see Appendix~\ref{app:main_results_extended}). For these reasons, we did not extend its evaluation to other source languages beyond English and Chinese.

The method \texttt{M2MS} conducts cross-lingual summarization for each target language independently, disregarding semantic coherence across languages. Consequently, it consistently achieves the highest ROUGE-2 scores but ranks lowest in coherence metrics (CometKiwi and BLASER 2.0 (C)). Conversely, \texttt{S\&T} ensures the best semantic coherence across target languages by directly translating the source language summary for each target language. However, this often results in significant degradation in similarity with the references for each target language, as measured by ROUGE-2, and, in many cases, even diminishes similarity to the reference summary for the source language, as measured by BLASER 2.0 (R). This indicates that the MT model introduced errors compromising summary quality. 

Our approaches (\texttt{PivotRR} and \texttt{NeutralRR}) do not significantly degrade ROUGE-2 scores compared to \texttt{M2MS} and notably achieve the highest similarity to the the reference summary for the source language. As expected, our methods also significantly improve semantic coherence across different target languages compared to \texttt{M2MS}. \texttt{NeutralRR} performs comparably to \texttt{PivotRR} in terms of similarity to the reference summaries, and consistently outperforms it in terms of semantic coherence across target languages. This was expected because \texttt{NeutralRR} treats all languages equally and aims for a set of summaries with high similarity. Conversely, \texttt{PivotRR} utilizes a fixed pivot summary and seeks candidates in each target language that closely resemble the pivot.

\subsection{Effect of Varying the Number of Candidates}
\label{sec:exp_num_candidates}

In this experiment, we investigate how the performance of our methods changes as we vary the number of candidates for re-ranking, using English as the source language. To vary the number of candidates generated by beam search multinomial sampling, we kept the number of beams per output sequence constant and equal to 5 and varied the number of output sequences. The results are in Figure~\ref{fig:results_multicand}, where we show the averages and standard deviations across the seven target languages.

Interestingly, increasing the number of candidates does not affect the similarity between the selected summaries and their respective references, as evaluated by ROUGE-2. In addition, it has a positive effect on the similarity between the selected summaries and the reference in the source language, as measured by BLASER 2.0 (R). We justify this observation by the hypothesis that a set of summaries with high similarity can serve as a reliable indicator of summary quality, since it is unlikely that the model generates the same false information in multiple languages. This was illustrated in the example in Table~\ref{tab:summary_examples} for \texttt{NeutralRR}. Finally, as more candidates are considered, computation time increases, yet so does the similarity of selected summaries, as evaluated by CometKiwi. Notably, this similarity increase is more significant for \texttt{NeutralRR}, which is not limited by maximizing similarity to a fixed pivot summary.

\section{Conclusion}
\label{sec:conclusion}
This work introduces multi-target cross-lingual summarization to address the challenge of achieving coherent summaries across multiple target languages. We propose two re-ranking approaches tailored to this task, which improve semantic coherence across languages compared to conventional beam search decoding, while still preserving similarity to the reference summaries. In particular, one of these methods eliminates the need for a pivot language, thus treating all languages equally and eliminating potential biases arising from pivot language selection. Furthermore, we extended the evaluation framework for cross-lingual summarization by including the assessment of semantic coherence across different target languages.

\section*{Limitations}
While we believe that our approach has merit, it is equally important to recognize its inherent limitations. First, we anticipate that as large language models continue to improve and become fluent in more languages, instructing the model to produce summaries with identical information for all target languages will eventually be sufficient to satisfy our semantic coherence constraint. However, due to the autoregressive nature of language models, the order in which summaries are decoded in different languages may introduce bias, which is not present in our approach. Second, the success of our re-ranking approaches depends on the quality of the sampled candidates. If all candidates are of low quality, or if they have poor semantic coherence across target languages, our approaches will inevitably fail. Investigating computationally efficient ways to incorporate the semantic coherence constraint directly at decoding time is an interesting research direction. Finally, our method introduces increased computational complexity compared to the usual beam search decoding.

In our experimental evaluation, we acknowledge it would be valuable to conduct further validation through more datasets and human evaluation. Using more datasets would be difficult as most cross-lingual summarization datasets are presented as pairs of source-language document and target-language summary, limiting applicability to our multi-target setting. Human evaluation was constrained by the need for a pool of polyglot speakers and a considerable sample size, which were not feasible. Nevertheless, we believe our contributions are valuable as they identify a gap in cross-lingual summarization and propose a principled method to address it, irrespective of the particular performance of the models implementing it.

\section*{Acknowledgements}
The authors thank André F.\ T.\ Martins (Unbabel, Instituto Superior Técnico, and Instituto de Telecomunicações) and João Gante (Hugging Face) for their insightful comments and suggestions, which were very helpful in shaping and implementing the ideas presented in this work. This research was supported by the Portuguese Recovery and Resilience Plan through project C645008882-00000055 (i.e., the Center For Responsible AI).

\bibliography{anthology,custom}

\begin{thebibliography}{42}
\expandafter\ifx\csname natexlab\endcsname\relax\def\natexlab#1{#1}\fi

\bibitem[{Artetxe and Schwenk(2019)}]{artetxe-schwenk-2019-massively}
Mikel Artetxe and Holger Schwenk. 2019.
\newblock \href {https://doi.org/10.1162/tacl_a_00288} {Massively multilingual sentence embeddings for zero-shot cross-lingual transfer and beyond}.
\newblock \emph{Transactions of the Association for Computational Linguistics}, 7:597--610.

\bibitem[{Bai et~al.(2021)Bai, Gao, and Huang}]{bai-etal-2021-cross}
Yu~Bai, Yang Gao, and Heyan Huang. 2021.
\newblock \href {https://doi.org/10.18653/v1/2021.acl-long.538} {Cross-lingual abstractive summarization with limited parallel resources}.
\newblock In \emph{Proceedings of the 59th Annual Meeting of the Association for Computational Linguistics and the 11th International Joint Conference on Natural Language Processing (Volume 1: Long Papers)}, pages 6910--6924, Online. Association for Computational Linguistics.

\bibitem[{Bhattacharjee et~al.(2023)Bhattacharjee, Hasan, Ahmad, Li, Kang, and Shahriyar}]{bhattacharjee-etal-2023-crosssum}
Abhik Bhattacharjee, Tahmid Hasan, Wasi~Uddin Ahmad, Yuan-Fang Li, Yong-Bin Kang, and Rifat Shahriyar. 2023.
\newblock \href {https://doi.org/10.18653/v1/2023.acl-long.143} {{C}ross{S}um: Beyond {E}nglish-centric cross-lingual summarization for 1,500+ language pairs}.
\newblock In \emph{Proceedings of the 61st Annual Meeting of the Association for Computational Linguistics (Volume 1: Long Papers)}, pages 2541--2564, Toronto, Canada. Association for Computational Linguistics.

\bibitem[{Cao et~al.(2020)Cao, Liu, and Wan}]{cao-etal-2020-jointly}
Yue Cao, Hui Liu, and Xiaojun Wan. 2020.
\newblock \href {https://doi.org/10.18653/v1/2020.acl-main.554} {Jointly learning to align and summarize for neural cross-lingual summarization}.
\newblock In \emph{Proceedings of the 58th Annual Meeting of the Association for Computational Linguistics}, pages 6220--6231, Online. Association for Computational Linguistics.

\bibitem[{Chen et~al.(2023{\natexlab{a}})Chen, Duquenne, Andrews, Kao, Mourachko, Schwenk, and Costa-juss{\`a}}]{chen-etal-2023-blaser}
Mingda Chen, Paul-Ambroise Duquenne, Pierre Andrews, Justine Kao, Alexandre Mourachko, Holger Schwenk, and Marta~R. Costa-juss{\`a}. 2023{\natexlab{a}}.
\newblock \href {https://doi.org/10.18653/v1/2023.acl-long.504} {{BLASER}: A text-free speech-to-speech translation evaluation metric}.
\newblock In \emph{Proceedings of the 61st Annual Meeting of the Association for Computational Linguistics (Volume 1: Long Papers)}, pages 9064--9079, Toronto, Canada. Association for Computational Linguistics.

\bibitem[{Chen et~al.(2023{\natexlab{b}})Chen, Zhang, Zhou, Bai, Wang, Zhong, Yan, Li, Li, Zhu, and Zhang}]{chen-etal-2023-revisiting}
Yulong Chen, Huajian Zhang, Yijie Zhou, Xuefeng Bai, Yueguan Wang, Ming Zhong, Jianhao Yan, Yafu Li, Judy Li, Xianchao Zhu, and Yue Zhang. 2023{\natexlab{b}}.
\newblock \href {https://doi.org/10.18653/v1/2023.acl-long.519} {Revisiting cross-lingual summarization: A corpus-based study and a new benchmark with improved annotation}.
\newblock In \emph{Proceedings of the 61st Annual Meeting of the Association for Computational Linguistics (Volume 1: Long Papers)}, pages 9332--9351, Toronto, Canada. Association for Computational Linguistics.

\bibitem[{Communication et~al.(2023)Communication, Barrault, Chung, Meglioli, Dale, Dong, Duquenne, Elsahar, Gong, Heffernan, Hoffman, Klaiber, Li, Licht, Maillard, Rakotoarison, Sadagopan, Wenzek, Ye, Akula, Chen, Hachem, Ellis, Gonzalez, Haaheim, Hansanti, Howes, Huang, Hwang, Inaguma, Jain, Kalbassi, Kallet, Kulikov, Lam, Li, Ma, Mavlyutov, Peloquin, Ramadan, Ramakrishnan, Sun, Tran, Tran, Tufanov, Vogeti, Wood, Yang, Yu, Andrews, Balioglu, Costa-jussà, Celebi, Elbayad, Gao, Guzmán, Kao, Lee, Mourachko, Pino, Popuri, Ropers, Saleem, Schwenk, Tomasello, Wang, Wang, and Wang}]{seamlessM4T}
Seamless Communication, Loïc Barrault, Yu-An Chung, Mariano~Cora Meglioli, David Dale, Ning Dong, Paul-Ambroise Duquenne, Hady Elsahar, Hongyu Gong, Kevin Heffernan, John Hoffman, Christopher Klaiber, Pengwei Li, Daniel Licht, Jean Maillard, Alice Rakotoarison, Kaushik~Ram Sadagopan, Guillaume Wenzek, Ethan Ye, Bapi Akula, Peng-Jen Chen, Naji~El Hachem, Brian Ellis, Gabriel~Mejia Gonzalez, Justin Haaheim, Prangthip Hansanti, Russ Howes, Bernie Huang, Min-Jae Hwang, Hirofumi Inaguma, Somya Jain, Elahe Kalbassi, Amanda Kallet, Ilia Kulikov, Janice Lam, Daniel Li, Xutai Ma, Ruslan Mavlyutov, Benjamin Peloquin, Mohamed Ramadan, Abinesh Ramakrishnan, Anna Sun, Kevin Tran, Tuan Tran, Igor Tufanov, Vish Vogeti, Carleigh Wood, Yilin Yang, Bokai Yu, Pierre Andrews, Can Balioglu, Marta~R. Costa-jussà, Onur Celebi, Maha Elbayad, Cynthia Gao, Francisco Guzmán, Justine Kao, Ann Lee, Alexandre Mourachko, Juan Pino, Sravya Popuri, Christophe Ropers, Safiyyah Saleem, Holger Schwenk, Paden Tomasello, Changhan Wang, Jeff
  Wang, and Skyler Wang. 2023.
\newblock \href {http://arxiv.org/abs/2308.11596} {Seamless{M4T}: Massively multilingual \& multimodal machine translation}.
\newblock \emph{arXiv preprint arXiv:2308.11596}.

\bibitem[{Conneau et~al.(2020)Conneau, Khandelwal, Goyal, Chaudhary, Wenzek, Guzm{\'a}n, Grave, Ott, Zettlemoyer, and Stoyanov}]{conneau-etal-2020-unsupervised}
Alexis Conneau, Kartikay Khandelwal, Naman Goyal, Vishrav Chaudhary, Guillaume Wenzek, Francisco Guzm{\'a}n, Edouard Grave, Myle Ott, Luke Zettlemoyer, and Veselin Stoyanov. 2020.
\newblock \href {https://doi.org/10.18653/v1/2020.acl-main.747} {Unsupervised cross-lingual representation learning at scale}.
\newblock In \emph{Proceedings of the 58th Annual Meeting of the Association for Computational Linguistics}, pages 8440--8451, Online. Association for Computational Linguistics.

\bibitem[{Costa-juss{\`a} et~al.(2022)Costa-juss{\`a}, Cross, {\c{C}}elebi, Elbayad, Heafield, Heffernan, Kalbassi, Lam, Licht, Maillard et~al.}]{costa2022no}
Marta~R Costa-juss{\`a}, James Cross, Onur {\c{C}}elebi, Maha Elbayad, Kenneth Heafield, Kevin Heffernan, Elahe Kalbassi, Janice Lam, Daniel Licht, Jean Maillard, et~al. 2022.
\newblock \href {https://arxiv.org/abs/2207.04672} {No language left behind: Scaling human-centered machine translation}.
\newblock \emph{arXiv preprint arXiv:2207.04672}.

\bibitem[{Duquenne et~al.(2023)Duquenne, Schwenk, and Sagot}]{duquenne2023sonar}
Paul-Ambroise Duquenne, Holger Schwenk, and Benoît Sagot. 2023.
\newblock \href {http://arxiv.org/abs/2308.11466} {{SONAR}: Sentence-level multimodal and language-agnostic representations}.
\newblock \emph{arXiv preprint arXiv:2308.11466}.

\bibitem[{Fadaee et~al.(2018)Fadaee, Bisazza, and Monz}]{fadaee-etal-2018-examining}
Marzieh Fadaee, Arianna Bisazza, and Christof Monz. 2018.
\newblock \href {https://aclanthology.org/L18-1148} {Examining the tip of the iceberg: A data set for idiom translation}.
\newblock In \emph{Proceedings of the Eleventh International Conference on Language Resources and Evaluation ({LREC} 2018)}, Miyazaki, Japan. European Language Resources Association (ELRA).

\bibitem[{Feremans et~al.(2003)Feremans, Labbé, and Laporte}]{feremans2003generalized}
Corinne Feremans, Martine Labbé, and Gilbert Laporte. 2003.
\newblock \href {https://doi.org/https://doi.org/10.1016/S0377-2217(02)00404-6} {Generalized network design problems}.
\newblock \emph{European Journal of Operational Research}, 148(1):1--13.

\bibitem[{Hasan et~al.(2021)Hasan, Bhattacharjee, Islam, Mubasshir, Li, Kang, Rahman, and Shahriyar}]{hasan-etal-2021-xl}
Tahmid Hasan, Abhik Bhattacharjee, Md.~Saiful Islam, Kazi Mubasshir, Yuan-Fang Li, Yong-Bin Kang, M.~Sohel Rahman, and Rifat Shahriyar. 2021.
\newblock \href {https://doi.org/10.18653/v1/2021.findings-acl.413} {{XL}-sum: Large-scale multilingual abstractive summarization for 44 languages}.
\newblock In \emph{Findings of the Association for Computational Linguistics: ACL-IJCNLP 2021}, pages 4693--4703, Online. Association for Computational Linguistics.

\bibitem[{Huot et~al.(2024)Huot, Maynez, Alberti, Amplayo, Agrawal, Fierro, Narayan, and Lapata}]{huot-etal-2024-mplan}
Fantine Huot, Joshua Maynez, Chris Alberti, Reinald~Kim Amplayo, Priyanka Agrawal, Constanza Fierro, Shashi Narayan, and Mirella Lapata. 2024.
\newblock \href {https://aclanthology.org/2024.eacl-long.131} {$\mu${PLAN}: Summarizing using a content plan as cross-lingual bridge}.
\newblock In \emph{Proceedings of the 18th Conference of the European Chapter of the Association for Computational Linguistics (Volume 1: Long Papers)}, pages 2146--2163, St. Julian{'}s, Malta. Association for Computational Linguistics.

\bibitem[{Jiang et~al.(2023)Jiang, Sablayrolles, Mensch, Bamford, Chaplot, de~las Casas, Bressand, Lengyel, Lample, Saulnier, Lavaud, Lachaux, Stock, Scao, Lavril, Wang, Lacroix, and Sayed}]{jiang2023mistral}
Albert~Q. Jiang, Alexandre Sablayrolles, Arthur Mensch, Chris Bamford, Devendra~Singh Chaplot, Diego de~las Casas, Florian Bressand, Gianna Lengyel, Guillaume Lample, Lucile Saulnier, Lélio~Renard Lavaud, Marie-Anne Lachaux, Pierre Stock, Teven~Le Scao, Thibaut Lavril, Thomas Wang, Timothée Lacroix, and William~El Sayed. 2023.
\newblock \href {http://arxiv.org/abs/2310.06825} {Mistral 7b}.
\newblock \emph{arXiv preprint arXiv:2310.06825}.

\bibitem[{Joulin et~al.(2016{\natexlab{a}})Joulin, Grave, Bojanowski, Douze, Jégou, and Mikolov}]{joulin2016fasttextzip}
Armand Joulin, Edouard Grave, Piotr Bojanowski, Matthijs Douze, Hérve Jégou, and Tomas Mikolov. 2016{\natexlab{a}}.
\newblock \href {http://arxiv.org/abs/1612.03651} {Fasttext.zip: Compressing text classification models}.
\newblock \emph{arXiv preprint arXiv:1612.03651}.

\bibitem[{Joulin et~al.(2016{\natexlab{b}})Joulin, Grave, Bojanowski, and Mikolov}]{joulin2016bag}
Armand Joulin, Edouard Grave, Piotr Bojanowski, and Tomas Mikolov. 2016{\natexlab{b}}.
\newblock \href {http://arxiv.org/abs/1607.01759} {Bag of tricks for efficient text classification}.
\newblock \emph{arXiv preprint arXiv:1607.01759}.

\bibitem[{Kepler et~al.(2019{\natexlab{a}})Kepler, Tr{\'e}nous, Treviso, Vera, G{\'o}is, Farajian, Lopes, and Martins}]{kepler-etal-2019-unbabels}
Fabio Kepler, Jonay Tr{\'e}nous, Marcos Treviso, Miguel Vera, Ant{\'o}nio G{\'o}is, M.~Amin Farajian, Ant{\'o}nio~V. Lopes, and Andr{\'e} F.~T. Martins. 2019{\natexlab{a}}.
\newblock \href {https://doi.org/10.18653/v1/W19-5406} {Unbabel{'}s participation in the {WMT}19 translation quality estimation shared task}.
\newblock In \emph{Proceedings of the Fourth Conference on Machine Translation (Volume 3: Shared Task Papers, Day 2)}, pages 78--84, Florence, Italy. Association for Computational Linguistics.

\bibitem[{Kepler et~al.(2019{\natexlab{b}})Kepler, Tr{\'e}nous, Treviso, Vera, and Martins}]{kepler-etal-2019-openkiwi}
Fabio Kepler, Jonay Tr{\'e}nous, Marcos Treviso, Miguel Vera, and Andr{\'e} F.~T. Martins. 2019{\natexlab{b}}.
\newblock \href {https://doi.org/10.18653/v1/P19-3020} {{O}pen{K}iwi: An open source framework for quality estimation}.
\newblock In \emph{Proceedings of the 57th Annual Meeting of the Association for Computational Linguistics: System Demonstrations}, pages 117--122, Florence, Italy. Association for Computational Linguistics.

\bibitem[{Ladhak et~al.(2020)Ladhak, Durmus, Cardie, and McKeown}]{ladhak-etal-2020-wikilingua}
Faisal Ladhak, Esin Durmus, Claire Cardie, and Kathleen McKeown. 2020.
\newblock \href {https://doi.org/10.18653/v1/2020.findings-emnlp.360} {{W}iki{L}ingua: A new benchmark dataset for cross-lingual abstractive summarization}.
\newblock In \emph{Findings of the Association for Computational Linguistics: EMNLP 2020}, pages 4034--4048, Online. Association for Computational Linguistics.

\bibitem[{Lee(2020)}]{lee-2020-two}
Dongjun Lee. 2020.
\newblock \href {https://aclanthology.org/2020.wmt-1.118} {Two-phase cross-lingual language model fine-tuning for machine translation quality estimation}.
\newblock In \emph{Proceedings of the Fifth Conference on Machine Translation}, pages 1024--1028, Online. Association for Computational Linguistics.

\bibitem[{Liang et~al.(2022)Liang, Meng, Zhou, Xu, Chen, Su, and Zhou}]{liang-etal-2022-variational}
Yunlong Liang, Fandong Meng, Chulun Zhou, Jinan Xu, Yufeng Chen, Jinsong Su, and Jie Zhou. 2022.
\newblock \href {https://doi.org/10.18653/v1/2022.acl-long.148} {A variational hierarchical model for neural cross-lingual summarization}.
\newblock In \emph{Proceedings of the 60th Annual Meeting of the Association for Computational Linguistics (Volume 1: Long Papers)}, pages 2088--2099, Dublin, Ireland. Association for Computational Linguistics.

\bibitem[{Lin(2004)}]{lin-2004-rouge}
Chin-Yew Lin. 2004.
\newblock \href {https://aclanthology.org/W04-1013} {{ROUGE}: A package for automatic evaluation of summaries}.
\newblock In \emph{Text Summarization Branches Out}, pages 74--81, Barcelona, Spain. Association for Computational Linguistics.

\bibitem[{Liu et~al.(2020)Liu, Gu, Goyal, Li, Edunov, Ghazvininejad, Lewis, and Zettlemoyer}]{liu-etal-2020-multilingual-denoising}
Yinhan Liu, Jiatao Gu, Naman Goyal, Xian Li, Sergey Edunov, Marjan Ghazvininejad, Mike Lewis, and Luke Zettlemoyer. 2020.
\newblock \href {https://doi.org/10.1162/tacl_a_00343} {Multilingual denoising pre-training for neural machine translation}.
\newblock \emph{Transactions of the Association for Computational Linguistics}, 8:726--742.

\bibitem[{Maynez et~al.(2020)Maynez, Narayan, Bohnet, and McDonald}]{maynez-etal-2020-faithfulness}
Joshua Maynez, Shashi Narayan, Bernd Bohnet, and Ryan McDonald. 2020.
\newblock \href {https://doi.org/10.18653/v1/2020.acl-main.173} {On faithfulness and factuality in abstractive summarization}.
\newblock In \emph{Proceedings of the 58th Annual Meeting of the Association for Computational Linguistics}, pages 1906--1919, Online. Association for Computational Linguistics.

\bibitem[{Perez-Beltrachini and Lapata(2021)}]{perez-beltrachini-lapata-2021-models}
Laura Perez-Beltrachini and Mirella Lapata. 2021.
\newblock \href {https://doi.org/10.18653/v1/2021.emnlp-main.742} {Models and datasets for cross-lingual summarisation}.
\newblock In \emph{Proceedings of the 2021 Conference on Empirical Methods in Natural Language Processing}, pages 9408--9423, Online and Punta Cana, Dominican Republic. Association for Computational Linguistics.

\bibitem[{Radford et~al.(2018)Radford, Narasimhan, Salimans, and Sutskever}]{radford2018improving}
Alec Radford, Karthik Narasimhan, Tim Salimans, and Ilya Sutskever. 2018.
\newblock \href {https://cdn.openai.com/research-covers/language-unsupervised/language_understanding_paper.pdf} {Improving language understanding by generative pre-training}.
\newblock \emph{Technical report, OpenAI}.

\bibitem[{Rei et~al.(2020)Rei, Stewart, Farinha, and Lavie}]{rei-etal-2020-comet}
Ricardo Rei, Craig Stewart, Ana~C Farinha, and Alon Lavie. 2020.
\newblock \href {https://doi.org/10.18653/v1/2020.emnlp-main.213} {{COMET}: A neural framework for {MT} evaluation}.
\newblock In \emph{Proceedings of the 2020 Conference on Empirical Methods in Natural Language Processing (EMNLP)}, pages 2685--2702, Online. Association for Computational Linguistics.

\bibitem[{Rei et~al.(2022)Rei, Treviso, Guerreiro, Zerva, Farinha, Maroti, C.~de Souza, Glushkova, Alves, Coheur, Lavie, and Martins}]{rei-etal-2022-cometkiwi}
Ricardo Rei, Marcos Treviso, Nuno~M. Guerreiro, Chrysoula Zerva, Ana~C Farinha, Christine Maroti, Jos{\'e}~G. C.~de Souza, Taisiya Glushkova, Duarte Alves, Luisa Coheur, Alon Lavie, and Andr{\'e} F.~T. Martins. 2022.
\newblock \href {https://aclanthology.org/2022.wmt-1.60} {{C}omet{K}iwi: {IST}-unbabel 2022 submission for the quality estimation shared task}.
\newblock In \emph{Proceedings of the Seventh Conference on Machine Translation (WMT)}, pages 634--645, Abu Dhabi, United Arab Emirates (Hybrid). Association for Computational Linguistics.

\bibitem[{Specia et~al.(2018)Specia, Scarton, and Paetzold}]{specia2018quality}
Lucia Specia, Carolina Scarton, and Gustavo~Henrique Paetzold. 2018.
\newblock \href {https://doi.org/10.1007/978-3-031-02168-8} {\emph{Quality estimation for machine translation}}, volume~11.
\newblock Springer.

\bibitem[{Specia et~al.(2013)Specia, Shah, de~Souza, and Cohn}]{specia-etal-2013-quest}
Lucia Specia, Kashif Shah, Jose~G.C. de~Souza, and Trevor Cohn. 2013.
\newblock \href {https://aclanthology.org/P13-4014} {{Q}u{E}st - a translation quality estimation framework}.
\newblock In \emph{Proceedings of the 51st Annual Meeting of the Association for Computational Linguistics: System Demonstrations}, pages 79--84, Sofia, Bulgaria. Association for Computational Linguistics.

\bibitem[{Tang et~al.(2021)Tang, Tran, Li, Chen, Goyal, Chaudhary, Gu, and Fan}]{tang-etal-2021-multilingual}
Yuqing Tang, Chau Tran, Xian Li, Peng-Jen Chen, Naman Goyal, Vishrav Chaudhary, Jiatao Gu, and Angela Fan. 2021.
\newblock \href {https://doi.org/10.18653/v1/2021.findings-acl.304} {Multilingual translation from denoising pre-training}.
\newblock In \emph{Findings of the Association for Computational Linguistics: ACL-IJCNLP 2021}, pages 3450--3466, Online. Association for Computational Linguistics.

\bibitem[{Team et~al.(2024)Team, Mesnard, Hardin, Dadashi, Bhupatiraju, Pathak, Sifre, Rivière, Kale, Love, Tafti, Hussenot, Chowdhery, Roberts, Barua, Botev, Castro-Ros, Slone, Héliou, Tacchetti, Bulanova, Paterson, Tsai, Shahriari, Lan, Choquette-Choo, Crepy, Cer, Ippolito, Reid, Buchatskaya, Ni, Noland, Yan, Tucker, Muraru, Rozhdestvenskiy, Michalewski, Tenney, Grishchenko, Austin, Keeling, Labanowski, Lespiau, Stanway, Brennan, Chen, Ferret, Chiu, Mao-Jones, Lee, Yu, Millican, Sjoesund, Lee, Dixon, Reid, Mikuła, Wirth, Sharman, Chinaev, Thain, Bachem, Chang, Wahltinez, Bailey, Michel, Yotov, Sessa, Chaabouni, Comanescu, Jana, Anil, McIlroy, Liu, Mullins, Smith, Borgeaud, Girgin, Douglas, Pandya, Shakeri, De, Klimenko, Hennigan, Feinberg, Stokowiec, hui Chen, Ahmed, Gong, Warkentin, Peran, Giang, Farabet, Vinyals, Dean, Kavukcuoglu, Hassabis, Ghahramani, Eck, Barral, Pereira, Collins, Joulin, Fiedel, Senter, Andreev, and Kenealy}]{gemmateam2024gemma}
Gemma Team, Thomas Mesnard, Cassidy Hardin, Robert Dadashi, Surya Bhupatiraju, Shreya Pathak, Laurent Sifre, Morgane Rivière, Mihir~Sanjay Kale, Juliette Love, Pouya Tafti, Léonard Hussenot, Aakanksha Chowdhery, Adam Roberts, Aditya Barua, Alex Botev, Alex Castro-Ros, Ambrose Slone, Amélie Héliou, Andrea Tacchetti, Anna Bulanova, Antonia Paterson, Beth Tsai, Bobak Shahriari, Charline~Le Lan, Christopher~A. Choquette-Choo, Clément Crepy, Daniel Cer, Daphne Ippolito, David Reid, Elena Buchatskaya, Eric Ni, Eric Noland, Geng Yan, George Tucker, George-Christian Muraru, Grigory Rozhdestvenskiy, Henryk Michalewski, Ian Tenney, Ivan Grishchenko, Jacob Austin, James Keeling, Jane Labanowski, Jean-Baptiste Lespiau, Jeff Stanway, Jenny Brennan, Jeremy Chen, Johan Ferret, Justin Chiu, Justin Mao-Jones, Katherine Lee, Kathy Yu, Katie Millican, Lars~Lowe Sjoesund, Lisa Lee, Lucas Dixon, Machel Reid, Maciej Mikuła, Mateo Wirth, Michael Sharman, Nikolai Chinaev, Nithum Thain, Olivier Bachem, Oscar Chang, Oscar
  Wahltinez, Paige Bailey, Paul Michel, Petko Yotov, Pier~Giuseppe Sessa, Rahma Chaabouni, Ramona Comanescu, Reena Jana, Rohan Anil, Ross McIlroy, Ruibo Liu, Ryan Mullins, Samuel~L Smith, Sebastian Borgeaud, Sertan Girgin, Sholto Douglas, Shree Pandya, Siamak Shakeri, Soham De, Ted Klimenko, Tom Hennigan, Vlad Feinberg, Wojciech Stokowiec, Yu~hui Chen, Zafarali Ahmed, Zhitao Gong, Tris Warkentin, Ludovic Peran, Minh Giang, Clément Farabet, Oriol Vinyals, Jeff Dean, Koray Kavukcuoglu, Demis Hassabis, Zoubin Ghahramani, Douglas Eck, Joelle Barral, Fernando Pereira, Eli Collins, Armand Joulin, Noah Fiedel, Evan Senter, Alek Andreev, and Kathleen Kenealy. 2024.
\newblock \href {http://arxiv.org/abs/2403.08295} {Gemma: Open models based on gemini research and technology}.
\newblock \emph{arXiv preprint arXiv:2403.08295}.

\bibitem[{Touvron et~al.(2023)Touvron, Martin, Stone, Albert, Almahairi, Babaei, Bashlykov, Batra, Bhargava, Bhosale, Bikel, Blecher, Ferrer, Chen, Cucurull, Esiobu, Fernandes, Fu, Fu, Fuller, Gao, Goswami, Goyal, Hartshorn, Hosseini, Hou, Inan, Kardas, Kerkez, Khabsa, Kloumann, Korenev, Koura, Lachaux, Lavril, Lee, Liskovich, Lu, Mao, Martinet, Mihaylov, Mishra, Molybog, Nie, Poulton, Reizenstein, Rungta, Saladi, Schelten, Silva, Smith, Subramanian, Tan, Tang, Taylor, Williams, Kuan, Xu, Yan, Zarov, Zhang, Fan, Kambadur, Narang, Rodriguez, Stojnic, Edunov, and Scialom}]{touvron2023llama}
Hugo Touvron, Louis Martin, Kevin Stone, Peter Albert, Amjad Almahairi, Yasmine Babaei, Nikolay Bashlykov, Soumya Batra, Prajjwal Bhargava, Shruti Bhosale, Dan Bikel, Lukas Blecher, Cristian~Canton Ferrer, Moya Chen, Guillem Cucurull, David Esiobu, Jude Fernandes, Jeremy Fu, Wenyin Fu, Brian Fuller, Cynthia Gao, Vedanuj Goswami, Naman Goyal, Anthony Hartshorn, Saghar Hosseini, Rui Hou, Hakan Inan, Marcin Kardas, Viktor Kerkez, Madian Khabsa, Isabel Kloumann, Artem Korenev, Punit~Singh Koura, Marie-Anne Lachaux, Thibaut Lavril, Jenya Lee, Diana Liskovich, Yinghai Lu, Yuning Mao, Xavier Martinet, Todor Mihaylov, Pushkar Mishra, Igor Molybog, Yixin Nie, Andrew Poulton, Jeremy Reizenstein, Rashi Rungta, Kalyan Saladi, Alan Schelten, Ruan Silva, Eric~Michael Smith, Ranjan Subramanian, Xiaoqing~Ellen Tan, Binh Tang, Ross Taylor, Adina Williams, Jian~Xiang Kuan, Puxin Xu, Zheng Yan, Iliyan Zarov, Yuchen Zhang, Angela Fan, Melanie Kambadur, Sharan Narang, Aurelien Rodriguez, Robert Stojnic, Sergey Edunov, and Thomas
  Scialom. 2023.
\newblock \href {http://arxiv.org/abs/2307.09288} {Llama 2: Open foundation and fine-tuned chat models}.
\newblock \emph{arXiv preprint arXiv:2307.09288}.

\bibitem[{Urlana et~al.(2023)Urlana, Chen, Zhao, Cohen, Shrivastava, and Haddow}]{urlana-etal-2023-pmindiasum}
Ashok Urlana, Pinzhen Chen, Zheng Zhao, Shay Cohen, Manish Shrivastava, and Barry Haddow. 2023.
\newblock \href {https://doi.org/10.18653/v1/2023.findings-emnlp.777} {{PMI}ndia{S}um: Multilingual and cross-lingual headline summarization for languages in {I}ndia}.
\newblock In \emph{Findings of the Association for Computational Linguistics: EMNLP 2023}, pages 11606--11628, Singapore. Association for Computational Linguistics.

\bibitem[{Vijayakumar et~al.(2018)Vijayakumar, Cogswell, Selvaraju, Sun, Lee, Crandall, and Batra}]{vijayakumar2018diverse}
Ashwin Vijayakumar, Michael Cogswell, Ramprasaath Selvaraju, Qing Sun, Stefan Lee, David Crandall, and Dhruv Batra. 2018.
\newblock \href {https://ojs.aaai.org/index.php/AAAI/article/view/12340} {Diverse beam search for improved description of complex scenes}.
\newblock In \emph{Proceedings of the AAAI Conference on Artificial Intelligence}, volume~32.

\bibitem[{Wang et~al.(2023{\natexlab{a}})Wang, Liang, Meng, Zou, Li, Qu, and Zhou}]{wang-etal-2023-zero}
Jiaan Wang, Yunlong Liang, Fandong Meng, Beiqi Zou, Zhixu Li, Jianfeng Qu, and Jie Zhou. 2023{\natexlab{a}}.
\newblock \href {https://doi.org/10.18653/v1/2023.newsum-1.2} {Zero-shot cross-lingual summarization via large language models}.
\newblock In \emph{Proceedings of the 4th New Frontiers in Summarization Workshop}, pages 12--23, Singapore. Association for Computational Linguistics.

\bibitem[{Wang et~al.(2022)Wang, Meng, Zheng, Liang, Li, Qu, and Zhou}]{wang-etal-2022-survey}
Jiaan Wang, Fandong Meng, Duo Zheng, Yunlong Liang, Zhixu Li, Jianfeng Qu, and Jie Zhou. 2022.
\newblock \href {https://doi.org/10.1162/tacl_a_00520} {A survey on cross-lingual summarization}.
\newblock \emph{Transactions of the Association for Computational Linguistics}, 10:1304--1323.

\bibitem[{Wang et~al.(2023{\natexlab{b}})Wang, Meng, Zheng, Liang, Li, Qu, and Zhou}]{wang-etal-2023-towards-unifying}
Jiaan Wang, Fandong Meng, Duo Zheng, Yunlong Liang, Zhixu Li, Jianfeng Qu, and Jie Zhou. 2023{\natexlab{b}}.
\newblock \href {https://doi.org/10.18653/v1/2023.acl-long.843} {Towards unifying multi-lingual and cross-lingual summarization}.
\newblock In \emph{Proceedings of the 61st Annual Meeting of the Association for Computational Linguistics (Volume 1: Long Papers)}, pages 15127--15143, Toronto, Canada. Association for Computational Linguistics.

\bibitem[{Wang et~al.(2018)Wang, Fan, Li, Zhou, Chen, Shi, and Si}]{wang-etal-2018-alibaba}
Jiayi Wang, Kai Fan, Bo~Li, Fengming Zhou, Boxing Chen, Yangbin Shi, and Luo Si. 2018.
\newblock \href {https://doi.org/10.18653/v1/W18-6465} {{A}libaba submission for {WMT}18 quality estimation task}.
\newblock In \emph{Proceedings of the Third Conference on Machine Translation: Shared Task Papers}, pages 809--815, Belgium, Brussels. Association for Computational Linguistics.

\bibitem[{Xue et~al.(2021)Xue, Constant, Roberts, Kale, Al-Rfou, Siddhant, Barua, and Raffel}]{xue-etal-2021-mt5}
Linting Xue, Noah Constant, Adam Roberts, Mihir Kale, Rami Al-Rfou, Aditya Siddhant, Aditya Barua, and Colin Raffel. 2021.
\newblock \href {https://doi.org/10.18653/v1/2021.naacl-main.41} {m{T}5: A massively multilingual pre-trained text-to-text transformer}.
\newblock In \emph{Proceedings of the 2021 Conference of the North American Chapter of the Association for Computational Linguistics: Human Language Technologies}, pages 483--498, Online. Association for Computational Linguistics.

\bibitem[{Zhu et~al.(2019)Zhu, Wang, Wang, Zhou, Zhang, Wang, and Zong}]{zhu-etal-2019-ncls}
Junnan Zhu, Qian Wang, Yining Wang, Yu~Zhou, Jiajun Zhang, Shaonan Wang, and Chengqing Zong. 2019.
\newblock \href {https://doi.org/10.18653/v1/D19-1302} {{NCLS}: Neural cross-lingual summarization}.
\newblock In \emph{Proceedings of the 2019 Conference on Empirical Methods in Natural Language Processing and the 9th International Joint Conference on Natural Language Processing (EMNLP-IJCNLP)}, pages 3054--3064, Hong Kong, China. Association for Computational Linguistics.

\end{thebibliography}
\bibliographystyle{acl_natbib}

\appendix

\section{Dataset Clustering and Analysis}
\label{app:dataset_analysis}

As mentioned in Section~\ref{sec:dataset}, the original CrossSum dataset presents documents in one language paired with summaries in another language, a format that does not serve our multi-target setting. Therefore, we clustered the dataset to obtain clusters of multilingual document-summary pairs about the same story. To achieve this, we aggregated all documents across the mentioned languages and constructed an undirected graph representing their pairwise connections. In this graph, two documents in different languages are connected if they are paired in CrossSum. We then built clusters by extracting all maximal cliques from this graph and we discarded all singleton cliques. Consequently, each maximal clique is a cluster of up to seven multilingual documents pertaining to the same story, where each document is accompanied by a summary in its respective language.

\begin{figure}[t]
\centering
\includegraphics[width=0.45\textwidth]{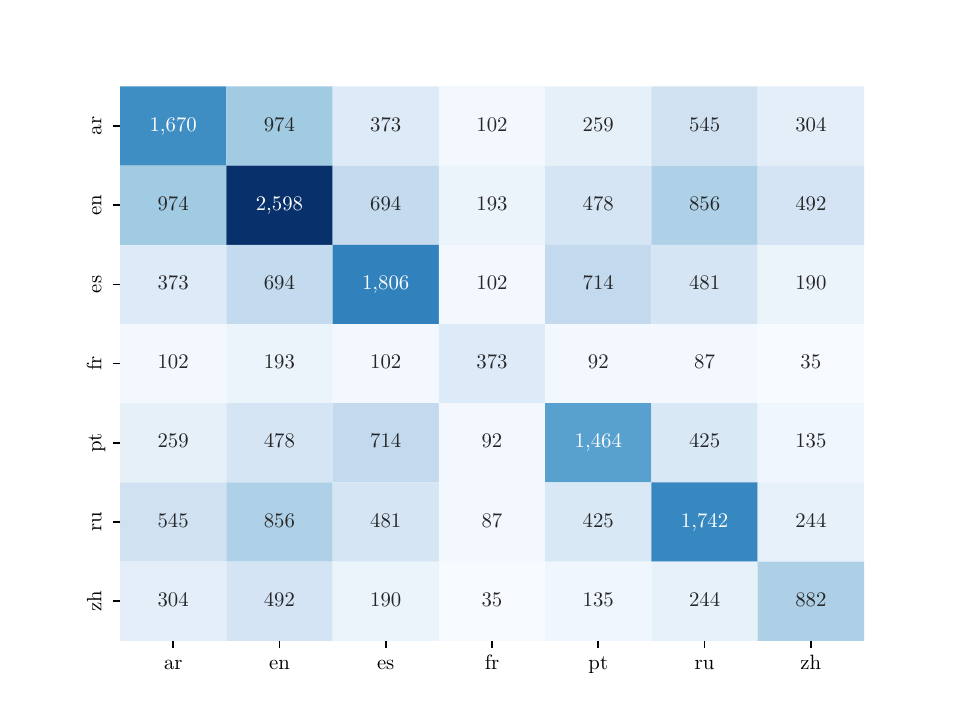}
    \caption{Number of clusters in the test set containing documents of each language pair.}\label{fig:crosssum_clustered_lang_pairs}
\end{figure}

This clustering procedure was applied separately to the CrossSum validation and test splits. The resulting validation set consisted of 4,525 clusters and 10,479 documents, while the test set consisted of 4,560 clusters and 10,535 documents. Table~\ref{tab:cluster_sizes} provides a breakdown of cluster sizes in the test set, as well as the distribution of documents for each language and cluster size. Notably, none of the clusters in the test set are complete, indicating that no cluster includes a document for all seven languages considered. In addition, we conducted an analysis of the co-occurrence of different language pairs within the clusters to verify whether a robust evaluation of cross-lingual summarization was possible across all language directions. Figure~\ref{fig:crosssum_clustered_lang_pairs} illustrates the distribution of clusters containing examples of each language pair. While certain language pairs have higher representation than others, it is noteworthy that even the least represented pair (fr, zh) is found in 35 clusters, indicating a diverse linguistic coverage across the dataset.

\begin{figure*}[t]
\centering
\begin{subfigure}[t]{0.45\textwidth}
  \centering
  \includegraphics[width=0.95\textwidth]{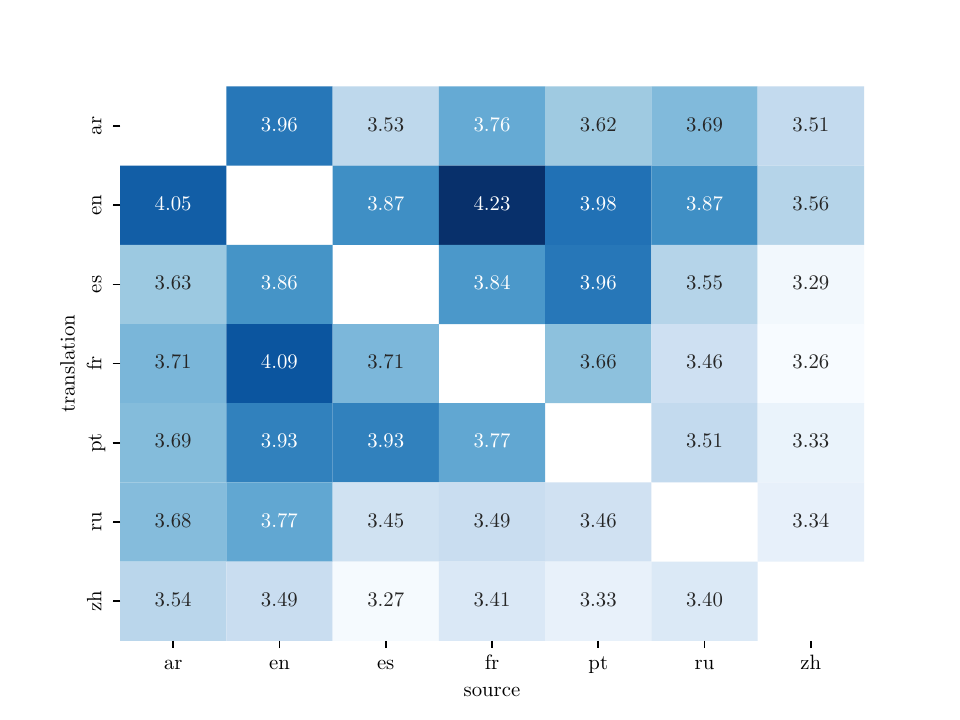}
  \caption{}
  \label{fig:blaser_crosssum}
\end{subfigure}
\begin{subfigure}[t]{0.45\textwidth}
  \centering
  \includegraphics[width=0.95\textwidth]{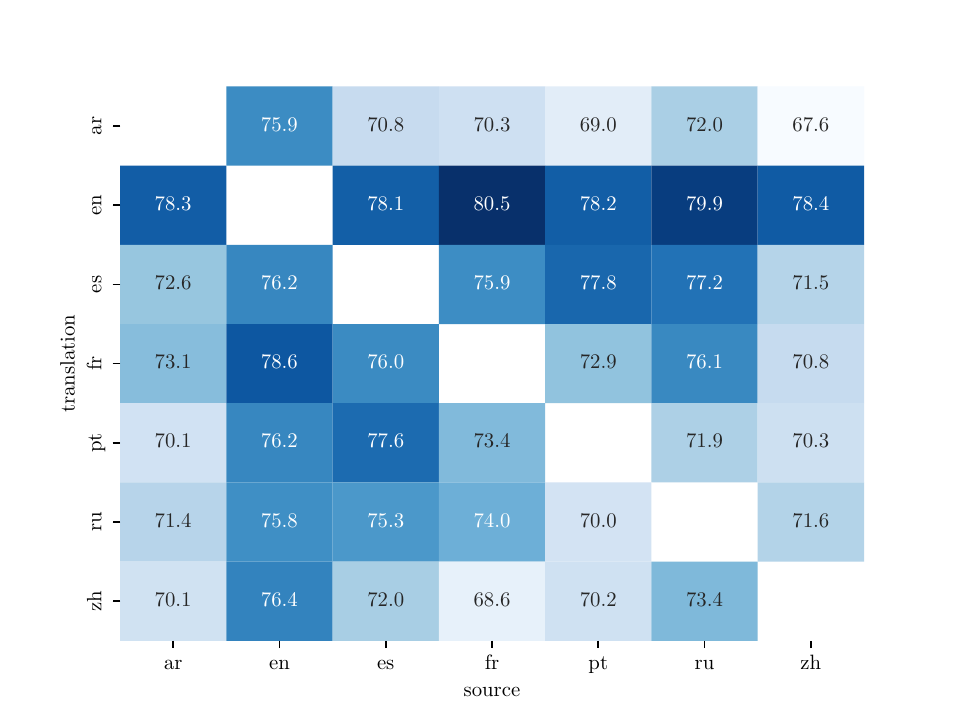}
  \caption{}
  \label{fig:cometkiwi_crosssum}
\end{subfigure}
\caption{Average BLASER 2.0 (a) and CometKiwi (b) scores between reference summaries within the same cluster for each language pair in the test set.}
\label{fig:crosssum_similarities}
\end{figure*}

Since one of our goals is to assess the semantic coherence of the generated summaries in different target languages, it is crucial to evaluate the coherence of reference summary clusters in this regard. This evaluation helps to determine the level of coherence that can be achieved in the generated summaries without degrading similarity to the reference summaries. To achieve this, we computed BLASER 2.0 and CometKiwi scores between reference summaries within the same cluster for each language pair. The results are shown in Figure~\ref{fig:crosssum_similarities}. It is important to note that the matrices are non-symmetric due to the nature of BLASER 2.0 and CometKiwi metrics. Firstly, we note a significant agreement between the two metrics, as anticipated. Additionally, coherence tends to be higher among languages using the Latin script. However, for most language pairs, coherence remains above 3.40 BLASER 2.0 points and 70.0 CometKiwi points. This suggests room for improvement compared to the results outlined in Table~\ref{tab:main_results}.

\begin{table}
\centering
\small
\begin{tabular}{lrrrrrrr} \toprule
\multirow{2}{*}{Language} & \multicolumn{7}{c}{Cluster Size} \\
                          & \multicolumn{1}{c}{2} & \multicolumn{1}{c}{3} & \multicolumn{1}{c}{4} & \multicolumn{1}{c}{5} & \multicolumn{1}{c}{6} & \multicolumn{1}{c}{7} & All \\ \midrule
ar                        & 1,022                 & 455                   & 153                   & 34                    & 6                     & 0    & 1,670                 \\
en                        & 1,780                 & 598                   & 176                   & 37                    & 7                     & 0    &  2,598                \\
es                        & 1,271                 & 367                   & 130                   & 31                    & 7                     & 0    &  1,806                \\
fr                        & 224                   & 84                    & 46                    & 14                    & 5                     & 0    &   373               \\
pt                        & 1,027                  & 280                   & 118                   & 33                    & 6                     & 0   &  1,464                \\
ru                        & 1,077                 & 482                   & 140                   & 38                    & 5                     & 0   & 1,742                  \\
zh                   & 531                   & 224                   & 93                    & 28                    & 6                     & 0  &  882                  \\ \midrule
All                       & 3,466                 & 830                   & 214                   & 43                    & 7                     & 0  &  4,560        \\ \bottomrule         
\end{tabular}
\caption{Number of clusters in the test set containing a document of each language, organized by cluster size.}
\label{tab:cluster_sizes}
\end{table}

\section{Implementation Details}
\label{app:imp_details}
To represent the cross-lingual summarization distribution $q(\boldsymbol{y}_t \mid \boldsymbol{x}_o, t)$, we use an mT5 model \cite{xue-etal-2021-mt5} for all the methods except \texttt{Mistral 7B}. mT5 allows us to perform summarization across all language directions by conditioning the decoder on a unique start-of-sequence token that specifies the intended target language.

We used the publicly available SONAR checkpoint \texttt{text\_sonar\_basic\_encoder} to implement $\phi$, the mT5 checkpoint \texttt{csebuetnlp/mT5\_m2m\_crossSum\_enhanced}, which was fine-tuned in the CrossSum dataset, and the Mistral 7B checkpoint \texttt{mistralai/Mistral-7B-Instruct-v0.2}. All of these checkpoints are available at the Hugging Face model hub.\footnote{\url{https://huggingface.co/models}}  

The optimal beam size and sampling temperature for beam search multinomial sampling were determined through a grid search. We explored beam sizes of 1, 3, and 5, and temperatures of 0.1, 0.3, 0.5, 1.0, 1.5, and 2.0 in order to maximize the ROUGE-2 score on the validation set of English-to-all summarization. We also tried with other decoding strategies, namely (single-beam) multinomial sampling and diverse beam search~\cite{vijayakumar2018diverse}, but these degraded ROUGE scores considerably. The number of random language permutations ($m$ in Algorithm~\ref{alg:multi_target_sum}) used by \texttt{NeutralRR} was set to 6 when the number of target languages was at least three and was set to 2 if there were only two target languages, since there are only two possible permutations of two languages.

Regarding the evaluation metrics, we used the multilingual implementation of ROUGE by \citet{hasan-etal-2021-xl}.\footnote{\url{https://github.com/csebuetnlp/xl-sum/tree/master/multilingual_rouge_scoring}} For CometKiwi and BLASER~2.0, we used the \texttt{Unbabel/wmt22-cometkiwi-da} and \texttt{blaser\_2\_0\_qe}  checkpoints, respectively.

All experiments were run on an 80-core Intel Xeon Gold 5218R CPU @ 2.10GHz with 800GB of RAM and an NVIDIA A100 GPU with 80GB of memory.

\section{LLM Prompt}
\label{app:llm_prompt}
The following prompt was used on the experiments with \texttt{Mistral 7B}:
\vspace{11pt}

\noindent\texttt{For the <source\_lang> news article from BBC written below, provide a summary in <target\_lang\_1>, a summary in <target\_lang\_2>, ... and a summary in <target\_lang\_N>. All summaries should be one or two sentences long and follow the style of BBC. All summaries must contain the same information. Present the answer in the format of a JSON object where the keys are the language codes and the values are the summaries.}

\noindent\texttt{Text:}

\noindent\texttt{<source\_document>}

\section{Further Experimental Results}
\label{app:further_exp_results}

\subsection{Main Results Extended}
\label{app:main_results_extended}

\begin{figure*}[t]
\centering
\includegraphics[width=1.0\textwidth]{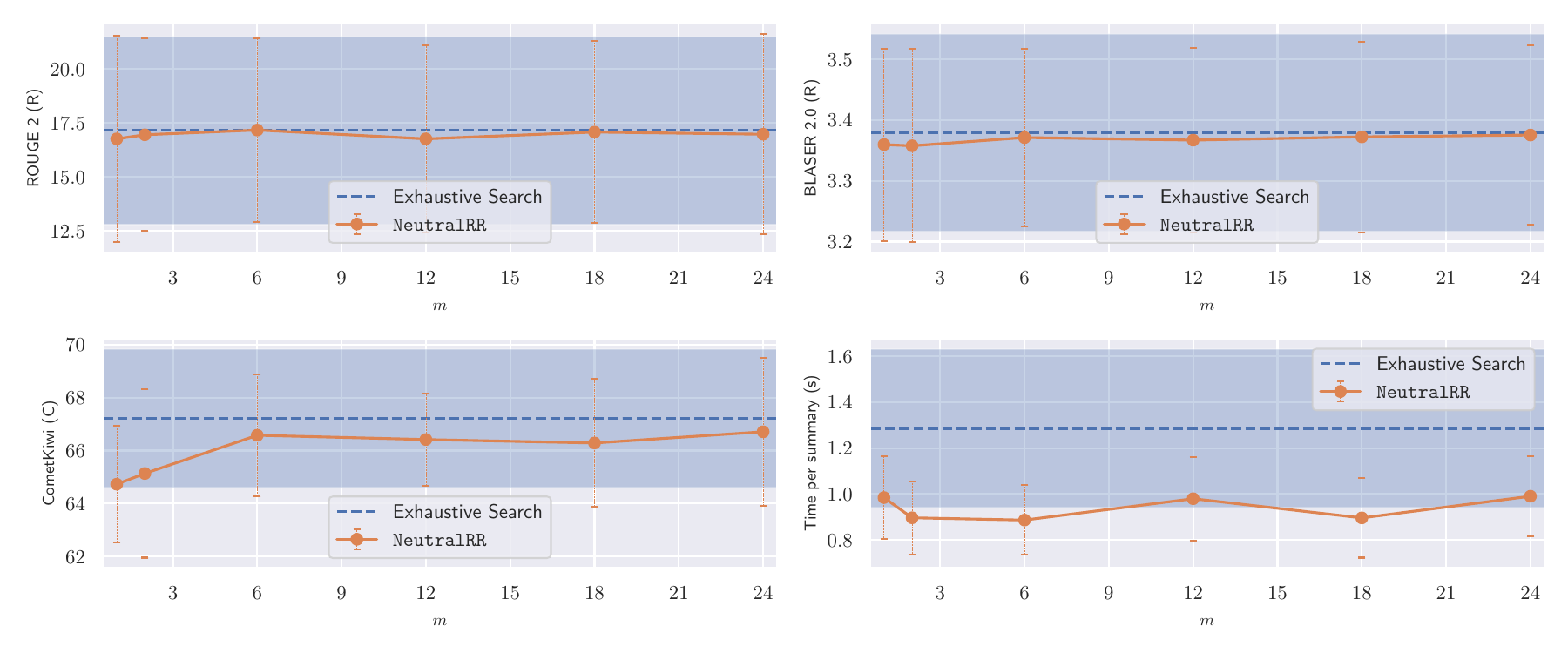}
    \caption{Effect of varying the number of language permutations ($m$ in Algorithm~\ref{alg:multi_target_sum}) on the results of \texttt{NeutralRR}. The results of doing an exhaustive search for the most coherent set are also shown for comparison. The error bars and the shaded area indicate the standard deviations across the target languages.}
    \label{fig:results_multiperm}
\end{figure*}

An extended version of the results presented in Table~\ref{tab:main_results} is shown in Tables~\ref{tab:extended_results_reference}~and~\ref{tab:extended_results_coherence}. In addition to English and Chinese, we also show results for Spanish and French. Spanish is the second most represented language in the dataset, surpassed only by English, while French is the least represented (see Table~\ref{tab:cluster_sizes}). All the results are accompanied by 95\% bootstrap confidence intervals with 1,000 resamples. Apart from the metrics mentioned in Section~\ref{sec:metrics}, we also include the target language accuracy in Table~\ref{tab:extended_results_reference}. This metric corresponds to the percentage of times a method generated text in the specified target language, and is calculated by comparing the specified language with the dominant language identified in the generated text by the fastText model \cite{joulin2016fasttextzip,joulin2016bag}. We observe that the mT5-based methods generate text in the correct target language in the vast majority (if not all) of the cases. \texttt{Mistral 7B} sometimes struggles to generate text in the correct target language, especially in Arabic.

\subsection{Effect of the Heuristic Search}
\label{app:exp_heuristic_search}

In this experiment, we investigate the effect of the number of language permutations ($m$ in Algorithm~\ref{alg:multi_target_sum}) on the performance of \texttt{NeutralRR}. In this experiment, we always use English as the source language and only consider clusters of documents with 4 languages, allowing up to 24 language permutations. The number of candidate summaries per language is kept fixed at 8. For this cluster size and number of candidates, maximizing $\varphi$ (equation~(\ref{eq:set_similarity})) directly with an exhaustive search is feasible since there are only $\text{8}^\text{4}\text{ = 4,096}$ possible sets of summaries. Therefore, we also compare the results of our approach with the exhaustive search. The results are shown in Figure~\ref{fig:results_multiperm}. 

The first observation is that changing $m$ or performing an exhaustive search does not significantly affect the similarity to the reference summaries. Changing $m$ also has no significant effect on the computation time, which is natural since the time required by the dynamic programming optimization is much smaller than the decoding time of the summarization model. However, an exhaustive search obviously increases the computation time, and the difference would only become larger for larger cluster sizes or more candidate summaries per language. Regarding the semantic coherence of the resulting set of summaries, an exhaustive search yields the best results as expected, but they are only slightly better than our heuristic search with a sufficiently large number of language permutations. 

\begin{landscape}
\begin{table}
\scriptsize
\centering
\begin{tabular}{llrrrrr|rrrrr|rrrrr}
\toprule
\multirow{2}{*}{Source} & \multirow{2}{*}{Method} & \multicolumn{5}{c}{ROUGE-2 (R)}                                                    & \multicolumn{5}{c}{BLASER 2.0 (R)} & \multicolumn{5}{c}{Target Lang. Acc.} \\
                        &                         & \multicolumn{1}{c}{en} & \multicolumn{1}{c}{es} & \multicolumn{1}{c}{fr} & \multicolumn{1}{c}{zh} & \multicolumn{1}{c}{rest} & \multicolumn{1}{c}{en} & \multicolumn{1}{c}{es} & \multicolumn{1}{c}{fr} & \multicolumn{1}{c}{zh} & \multicolumn{1}{c}{rest} & \multicolumn{1}{c}{en} & \multicolumn{1}{c}{es} & \multicolumn{1}{c}{fr} & \multicolumn{1}{c}{zh} & \multicolumn{1}{c}{rest} \\ \midrule
\multirow{5}{*}{en} & \texttt{M2MS} & 17.88 {\tiny\color{gray}$\pm$ 0.59} & 13.91 {\tiny\color{gray}$\pm$ 0.97} & 21.55 {\tiny\color{gray}$\pm$ 3.14} & 18.63 {\tiny\color{gray}$\pm$ 1.82} & 10.19 {\tiny\color{gray}$\pm$ 0.83} & 3.52 {\tiny\color{gray}$\pm$ 0.02} & 3.03 {\tiny\color{gray}$\pm$ 0.03} & 3.31 {\tiny\color{gray}$\pm$ 0.08} & 3.04 {\tiny\color{gray}$\pm$ 0.05} & 3.31 {\tiny\color{gray}$\pm$ 0.03} & 100.0 & 99.7 & 100.0 & 97.8 & 99.6\\
           & \texttt{S\&T} & 17.88 {\tiny\color{gray}$\pm$ 0.59} & 11.91 {\tiny\color{gray}$\pm$ 0.84} & 18.98 {\tiny\color{gray}$\pm$ 2.52} & 7.51 {\tiny\color{gray}$\pm$ 0.78} & 9.33 {\tiny\color{gray}$\pm$ 0.77} & 3.52 {\tiny\color{gray}$\pm$ 0.02} & 3.07 {\tiny\color{gray}$\pm$ 0.04} & 3.30 {\tiny\color{gray}$\pm$ 0.07} & 2.73 {\tiny\color{gray}$\pm$ 0.04} & 3.31 {\tiny\color{gray}$\pm$ 0.04} & 100.0 & 99.9 & 100.0 & 99.8 & 99.9\\
           & \texttt{Mistral 7B} & 6.52 {\tiny\color{gray}$\pm$ 0.28} & 6.68 {\tiny\color{gray}$\pm$ 0.58} & 7.86 {\tiny\color{gray}$\pm$ 1.57} & 3.18 {\tiny\color{gray}$\pm$ 0.37} & 2.78 {\tiny\color{gray}$\pm$ 0.29} & 2.45 {\tiny\color{gray}$\pm$ 0.02} & 2.17 {\tiny\color{gray}$\pm$ 0.03} & 2.46 {\tiny\color{gray}$\pm$ 0.08} & 2.13 {\tiny\color{gray}$\pm$ 0.03} & 2.31 {\tiny\color{gray}$\pm$ 0.03} & 99.7 & 99.8 & 100.0 & 99.2 & 82.2\\
           & \texttt{PivotRR} & 17.88 {\tiny\color{gray}$\pm$ 0.59} & 13.58 {\tiny\color{gray}$\pm$ 0.96} & 20.77 {\tiny\color{gray}$\pm$ 3.08} & 17.54 {\tiny\color{gray}$\pm$ 1.69} & 10.42 {\tiny\color{gray}$\pm$ 0.85} & 3.52 {\tiny\color{gray}$\pm$ 0.02} & 3.06 {\tiny\color{gray}$\pm$ 0.03} & 3.32 {\tiny\color{gray}$\pm$ 0.08} & 3.09 {\tiny\color{gray}$\pm$ 0.04} & 3.34 {\tiny\color{gray}$\pm$ 0.03} & 100.0 & 99.4 & 100.0 & 98.8 & 99.6\\
           & \texttt{NeutralRR} & 17.59 {\tiny\color{gray}$\pm$ 0.58} & 13.70 {\tiny\color{gray}$\pm$ 0.94} & 20.08 {\tiny\color{gray}$\pm$ 3.24} & 17.87 {\tiny\color{gray}$\pm$ 1.75} & 10.23 {\tiny\color{gray}$\pm$ 0.86} & 3.53 {\tiny\color{gray}$\pm$ 0.02} & 3.06 {\tiny\color{gray}$\pm$ 0.03} & 3.34 {\tiny\color{gray}$\pm$ 0.08} & 3.08 {\tiny\color{gray}$\pm$ 0.04} & 3.34 {\tiny\color{gray}$\pm$ 0.03} & 100.0 & 99.7 & 99.5 & 99.4 & 99.5\\
\midrule
\multirow{4}{*}{es} & \texttt{M2MS} & 15.88 {\tiny\color{gray}$\pm$ 1.26} & 14.37 {\tiny\color{gray}$\pm$ 0.66} & 20.48 {\tiny\color{gray}$\pm$ 4.77} & 20.01 {\tiny\color{gray}$\pm$ 3.38} & 9.12 {\tiny\color{gray}$\pm$ 1.04} & 3.40 {\tiny\color{gray}$\pm$ 0.04} & 2.99 {\tiny\color{gray}$\pm$ 0.02} & 3.26 {\tiny\color{gray}$\pm$ 0.15} & 2.92 {\tiny\color{gray}$\pm$ 0.10} & 3.20 {\tiny\color{gray}$\pm$ 0.04} & 99.7 & 99.8 & 100.0 & 97.9 & 99.9\\
           & \texttt{S\&T} & 10.71 {\tiny\color{gray}$\pm$ 0.80} & 14.37 {\tiny\color{gray}$\pm$ 0.66} & 14.28 {\tiny\color{gray}$\pm$ 2.91} & 7.99 {\tiny\color{gray}$\pm$ 1.31} & 8.01 {\tiny\color{gray}$\pm$ 0.85} & 3.29 {\tiny\color{gray}$\pm$ 0.04} & 2.99 {\tiny\color{gray}$\pm$ 0.02} & 3.16 {\tiny\color{gray}$\pm$ 0.12} & 2.60 {\tiny\color{gray}$\pm$ 0.07} & 3.15 {\tiny\color{gray}$\pm$ 0.04} & 100.0 & 99.8 & 100.0 & 99.5 & 99.8\\
           & \texttt{PivotRR} & 15.35 {\tiny\color{gray}$\pm$ 1.20} & 14.37 {\tiny\color{gray}$\pm$ 0.66} & 18.70 {\tiny\color{gray}$\pm$ 4.55} & 19.72 {\tiny\color{gray}$\pm$ 3.03} & 9.37 {\tiny\color{gray}$\pm$ 1.02} & 3.42 {\tiny\color{gray}$\pm$ 0.04} & 2.99 {\tiny\color{gray}$\pm$ 0.02} & 3.27 {\tiny\color{gray}$\pm$ 0.14} & 2.95 {\tiny\color{gray}$\pm$ 0.09} & 3.23 {\tiny\color{gray}$\pm$ 0.04} & 99.7 & 99.8 & 99.0 & 99.5 & 99.7\\
           & \texttt{NeutralRR} & 15.59 {\tiny\color{gray}$\pm$ 1.25} & 14.68 {\tiny\color{gray}$\pm$ 0.65} & 20.68 {\tiny\color{gray}$\pm$ 4.61} & 19.69 {\tiny\color{gray}$\pm$ 3.07} & 9.54 {\tiny\color{gray}$\pm$ 1.07} & 3.43 {\tiny\color{gray}$\pm$ 0.04} & 3.05 {\tiny\color{gray}$\pm$ 0.02} & 3.33 {\tiny\color{gray}$\pm$ 0.14} & 2.98 {\tiny\color{gray}$\pm$ 0.08} & 3.23 {\tiny\color{gray}$\pm$ 0.04} & 99.7 & 99.9 & 99.0 & 100.0 & 99.7\\
\midrule
\multirow{4}{*}{fr} & \texttt{M2MS} & 21.25 {\tiny\color{gray}$\pm$ 3.19} & 15.49 {\tiny\color{gray}$\pm$ 3.49} & 23.78 {\tiny\color{gray}$\pm$ 2.46} & 38.04 {\tiny\color{gray}$\pm$ 10.21} & 13.38 {\tiny\color{gray}$\pm$ 3.65} & 3.57 {\tiny\color{gray}$\pm$ 0.08} & 3.05 {\tiny\color{gray}$\pm$ 0.10} & 3.33 {\tiny\color{gray}$\pm$ 0.07} & 3.24 {\tiny\color{gray}$\pm$ 0.24} & 3.29 {\tiny\color{gray}$\pm$ 0.11} & 100.0 & 100.0 & 100.0 & 100.0 & 100.0\\
           & \texttt{S\&T} & 16.14 {\tiny\color{gray}$\pm$ 2.08} & 13.19 {\tiny\color{gray}$\pm$ 2.56} & 23.78 {\tiny\color{gray}$\pm$ 2.46} & 9.87 {\tiny\color{gray}$\pm$ 3.11} & 11.01 {\tiny\color{gray}$\pm$ 2.54} & 3.43 {\tiny\color{gray}$\pm$ 0.08} & 3.11 {\tiny\color{gray}$\pm$ 0.11} & 3.33 {\tiny\color{gray}$\pm$ 0.07} & 2.78 {\tiny\color{gray}$\pm$ 0.18} & 3.28 {\tiny\color{gray}$\pm$ 0.10} & 100.0 & 100.0 & 100.0 & 100.0 & 100.0\\
           & \texttt{PivotRR} & 20.11 {\tiny\color{gray}$\pm$ 2.98} & 15.31 {\tiny\color{gray}$\pm$ 3.00} & 23.78 {\tiny\color{gray}$\pm$ 2.46} & 38.21 {\tiny\color{gray}$\pm$ 9.44} & 13.16 {\tiny\color{gray}$\pm$ 3.33} & 3.54 {\tiny\color{gray}$\pm$ 0.08} & 3.10 {\tiny\color{gray}$\pm$ 0.11} & 3.33 {\tiny\color{gray}$\pm$ 0.07} & 3.35 {\tiny\color{gray}$\pm$ 0.21} & 3.34 {\tiny\color{gray}$\pm$ 0.11} & 100.0 & 100.0 & 100.0 & 100.0 & 100.0\\
           & \texttt{NeutralRR} & 20.95 {\tiny\color{gray}$\pm$ 2.79} & 14.42 {\tiny\color{gray}$\pm$ 3.13} & 22.76 {\tiny\color{gray}$\pm$ 2.41} & 36.28 {\tiny\color{gray}$\pm$ 10.27} & 13.18 {\tiny\color{gray}$\pm$ 3.29} & 3.59 {\tiny\color{gray}$\pm$ 0.08} & 3.08 {\tiny\color{gray}$\pm$ 0.11} & 3.33 {\tiny\color{gray}$\pm$ 0.06} & 3.26 {\tiny\color{gray}$\pm$ 0.21} & 3.36 {\tiny\color{gray}$\pm$ 0.11} & 100.0 & 100.0 & 100.0 & 100.0 & 100.0\\
\midrule
\multirow{5}{*}{zh} & \texttt{M2MS} & 17.95 {\tiny\color{gray}$\pm$ 1.52} & 15.54 {\tiny\color{gray}$\pm$ 2.24} & 30.91 {\tiny\color{gray}$\pm$ 10.19} & 24.13 {\tiny\color{gray}$\pm$ 1.54} & 11.72 {\tiny\color{gray}$\pm$ 1.77} & 3.58 {\tiny\color{gray}$\pm$ 0.05} & 3.04 {\tiny\color{gray}$\pm$ 0.07} & 3.50 {\tiny\color{gray}$\pm$ 0.27} & 3.14 {\tiny\color{gray}$\pm$ 0.04} & 3.34 {\tiny\color{gray}$\pm$ 0.06} & 98.0 & 100.0 & 100.0 & 99.8 & 99.6\\
           & \texttt{S\&T} & 13.51 {\tiny\color{gray}$\pm$ 1.17} & 11.36 {\tiny\color{gray}$\pm$ 1.42} & 21.79 {\tiny\color{gray}$\pm$ 5.13} & 24.13 {\tiny\color{gray}$\pm$ 1.54} & 9.13 {\tiny\color{gray}$\pm$ 1.41} & 3.48 {\tiny\color{gray}$\pm$ 0.04} & 3.00 {\tiny\color{gray}$\pm$ 0.08} & 3.32 {\tiny\color{gray}$\pm$ 0.16} & 3.14 {\tiny\color{gray}$\pm$ 0.04} & 3.31 {\tiny\color{gray}$\pm$ 0.07} & 100.0 & 99.5 & 100.0 & 99.8 & 99.3\\
           & \texttt{Mistral 7B} & 4.58 {\tiny\color{gray}$\pm$ 0.43} & 4.39 {\tiny\color{gray}$\pm$ 0.65} & 8.22 {\tiny\color{gray}$\pm$ 3.00} & 3.93 {\tiny\color{gray}$\pm$ 0.30} & 1.92 {\tiny\color{gray}$\pm$ 0.45} & 2.47 {\tiny\color{gray}$\pm$ 0.04} & 2.26 {\tiny\color{gray}$\pm$ 0.05} & 2.44 {\tiny\color{gray}$\pm$ 0.15} & 2.02 {\tiny\color{gray}$\pm$ 0.03} & 2.43 {\tiny\color{gray}$\pm$ 0.05} & 99.8 & 91.6 & 100.0 & 97.4 & 63.1\\
           & \texttt{PivotRR} & 18.32 {\tiny\color{gray}$\pm$ 1.46} & 15.73 {\tiny\color{gray}$\pm$ 2.09} & 30.67 {\tiny\color{gray}$\pm$ 8.23} & 24.13 {\tiny\color{gray}$\pm$ 1.54} & 11.80 {\tiny\color{gray}$\pm$ 1.65} & 3.60 {\tiny\color{gray}$\pm$ 0.05} & 3.09 {\tiny\color{gray}$\pm$ 0.08} & 3.51 {\tiny\color{gray}$\pm$ 0.18} & 3.14 {\tiny\color{gray}$\pm$ 0.04} & 3.39 {\tiny\color{gray}$\pm$ 0.06} & 98.4 & 100.0 & 100.0 & 99.8 & 99.9\\
           & \texttt{NeutralRR} & 18.34 {\tiny\color{gray}$\pm$ 1.46} & 15.63 {\tiny\color{gray}$\pm$ 2.11} & 29.25 {\tiny\color{gray}$\pm$ 7.83} & 23.72 {\tiny\color{gray}$\pm$ 1.47} & 12.32 {\tiny\color{gray}$\pm$ 1.73} & 3.61 {\tiny\color{gray}$\pm$ 0.05} & 3.09 {\tiny\color{gray}$\pm$ 0.07} & 3.50 {\tiny\color{gray}$\pm$ 0.20} & 3.18 {\tiny\color{gray}$\pm$ 0.04} & 3.39 {\tiny\color{gray}$\pm$ 0.06} & 98.6 & 100.0 & 100.0 & 99.9 & 99.9\\
\midrule
\multirow{4}{*}{rest} & \texttt{M2MS} & 15.50 {\tiny\color{gray}$\pm$ 1.15} & 12.82 {\tiny\color{gray}$\pm$ 1.10} & 22.16 {\tiny\color{gray}$\pm$ 5.59} & 20.36 {\tiny\color{gray}$\pm$ 3.16} & 11.02 {\tiny\color{gray}$\pm$ 0.99} & 3.48 {\tiny\color{gray}$\pm$ 0.04} & 2.99 {\tiny\color{gray}$\pm$ 0.04} & 3.29 {\tiny\color{gray}$\pm$ 0.14} & 3.06 {\tiny\color{gray}$\pm$ 0.08} & 3.30 {\tiny\color{gray}$\pm$ 0.04} & 99.9 & 99.4 & 99.7 & 99.6 & 99.9 \\
           & \texttt{S\&T} & 10.84 {\tiny\color{gray}$\pm$ 0.81} & 10.95 {\tiny\color{gray}$\pm$ 0.90} & 14.69 {\tiny\color{gray}$\pm$ 2.68} & 6.76 {\tiny\color{gray}$\pm$ 1.15} & 9.52 {\tiny\color{gray}$\pm$ 0.80} & 3.38 {\tiny\color{gray}$\pm$ 0.03} & 3.01 {\tiny\color{gray}$\pm$ 0.04} & 3.19 {\tiny\color{gray}$\pm$ 0.10} & 2.74 {\tiny\color{gray}$\pm$ 0.06} & 3.28 {\tiny\color{gray}$\pm$ 0.04} & 100.0 & 99.8 & 100.0 & 100.0 & 99.8 \\
           & \texttt{PivotRR} & 15.38 {\tiny\color{gray}$\pm$ 1.09} & 12.93 {\tiny\color{gray}$\pm$ 1.08} & 20.91 {\tiny\color{gray}$\pm$ 4.98} & 19.96 {\tiny\color{gray}$\pm$ 3.08} & 11.13 {\tiny\color{gray}$\pm$ 0.98} & 3.51 {\tiny\color{gray}$\pm$ 0.04} & 3.03 {\tiny\color{gray}$\pm$ 0.04} & 3.32 {\tiny\color{gray}$\pm$ 0.13} & 3.10 {\tiny\color{gray}$\pm$ 0.08} & 3.32 {\tiny\color{gray}$\pm$ 0.04} & 99.9 & 99.4 & 99.6 & 99.8 & 99.9 \\
           & \texttt{NeutralRR} & 15.29 {\tiny\color{gray}$\pm$ 1.13} & 13.04 {\tiny\color{gray}$\pm$ 1.09} & 21.14 {\tiny\color{gray}$\pm$ 4.78} & 19.70 {\tiny\color{gray}$\pm$ 2.94} & 11.18 {\tiny\color{gray}$\pm$ 0.99} & 3.50 {\tiny\color{gray}$\pm$ 0.04} & 3.03 {\tiny\color{gray}$\pm$ 0.04} & 3.34 {\tiny\color{gray}$\pm$ 0.13} & 3.11 {\tiny\color{gray}$\pm$ 0.08} & 3.34 {\tiny\color{gray}$\pm$ 0.04} & 99.8 & 99.5 & 100.0 & 99.6 & 100.0 \\
\bottomrule
\end{tabular}
\caption{Extended results of multi-target cross-lingual summarization in CrossSum for the metrics evaluating similarity to the reference summaries. The target language accuracy is also shown.}
\label{tab:extended_results_reference}
\end{table}
\end{landscape}

\begin{landscape}
\begin{table}
\scriptsize
\centering
\begin{tabular}{llrrrrr|rrrrr}
\toprule
\multirow{2}{*}{Source} & \multirow{2}{*}{Method} & \multicolumn{5}{c}{CometKiwi (C)}                                                    & \multicolumn{5}{c}{BLASER 2.0 (C)} \\
                        &                         & \multicolumn{1}{c}{en} & \multicolumn{1}{c}{es} & \multicolumn{1}{c}{fr} & \multicolumn{1}{c}{zh} & \multicolumn{1}{c}{rest} & \multicolumn{1}{c}{en}  & \multicolumn{1}{c}{es} & \multicolumn{1}{c}{fr} & \multicolumn{1}{c}{zh} & \multicolumn{1}{c}{rest} \\ \midrule
\multirow{5}{*}{en} & \texttt{M2MS} & 59.28 {\tiny\color{gray}$\pm$ 0.56} & 61.79 {\tiny\color{gray}$\pm$ 1.16} & 58.21 {\tiny\color{gray}$\pm$ 2.39} & 61.00 {\tiny\color{gray}$\pm$ 1.40} & 60.52 {\tiny\color{gray}$\pm$ 1.06} & 3.48 {\tiny\color{gray}$\pm$ 0.02} & 3.52 {\tiny\color{gray}$\pm$ 0.04} & 3.63 {\tiny\color{gray}$\pm$ 0.08} & 3.26 {\tiny\color{gray}$\pm$ 0.01} & 3.6 {\tiny\color{gray}$\pm$ 0.01} \\
           & \texttt{S\&T} & 85.00 {\tiny\color{gray}$\pm$ 0.20} & 87.33 {\tiny\color{gray}$\pm$ 0.33} & 87.93 {\tiny\color{gray}$\pm$ 0.39} & 79.24 {\tiny\color{gray}$\pm$ 1.15} & 85.58 {\tiny\color{gray}$\pm$ 0.39} & 4.67 {\tiny\color{gray}$\pm$ 0.01} & 4.80 {\tiny\color{gray}$\pm$ 0.02} & 4.81 {\tiny\color{gray}$\pm$ 0.03} & 3.87 {\tiny\color{gray}$\pm$ 0.05} & 4.78 {\tiny\color{gray}$\pm$ 0.03} \\
           & \texttt{Mistral 7B} & 69.77 {\tiny\color{gray}$\pm$ 0.66} & 79.40 {\tiny\color{gray}$\pm$ 0.75} & 77.97 {\tiny\color{gray}$\pm$ 2.01} & 65.95 {\tiny\color{gray}$\pm$ 1.04} & 66.04 {\tiny\color{gray}$\pm$ 0.96} & 3.09 {\tiny\color{gray}$\pm$ 0.04} & 3.44 {\tiny\color{gray}$\pm$ 0.08} & 3.19 {\tiny\color{gray}$\pm$ 0.15} & 3.24 {\tiny\color{gray}$\pm$ 0.07} & 3.06 {\tiny\color{gray}$\pm$ 0.06} \\
           & \texttt{PivotRR} & 63.72 {\tiny\color{gray}$\pm$ 0.52} & 64.79 {\tiny\color{gray}$\pm$ 1.15} & 61.51 {\tiny\color{gray}$\pm$ 2.48} & 64.42 {\tiny\color{gray}$\pm$ 1.28} & 63.49 {\tiny\color{gray}$\pm$ 1.03} & 3.71 {\tiny\color{gray}$\pm$ 0.02} & 3.73 {\tiny\color{gray}$\pm$ 0.03} & 3.83 {\tiny\color{gray}$\pm$ 0.06} & 3.45 {\tiny\color{gray}$\pm$ 0.04} & 3.83 {\tiny\color{gray}$\pm$ 0.03} \\
           & \texttt{NeutralRR} & 64.34 {\tiny\color{gray}$\pm$ 0.54} & 66.12 {\tiny\color{gray}$\pm$ 1.21} & 63.20 {\tiny\color{gray}$\pm$ 2.33} & 65.43 {\tiny\color{gray}$\pm$ 1.25} & 64.83 {\tiny\color{gray}$\pm$ 1.04} & 3.76 {\tiny\color{gray}$\pm$ 0.02} & 3.81 {\tiny\color{gray}$\pm$ 0.03} & 3.91 {\tiny\color{gray}$\pm$ 0.05} & 3.49 {\tiny\color{gray}$\pm$ 0.04} & 3.91 {\tiny\color{gray}$\pm$ 0.03} \\
\midrule
\multirow{4}{*}{es} & \texttt{M2MS} & 61.25 {\tiny\color{gray}$\pm$ 1.06} & 61.61 {\tiny\color{gray}$\pm$ 0.72} & 60.40 {\tiny\color{gray}$\pm$ 3.15} & 59.23 {\tiny\color{gray}$\pm$ 2.23} & 61.65 {\tiny\color{gray}$\pm$ 1.39} & 3.52 {\tiny\color{gray}$\pm$ 0.04} & 3.44 {\tiny\color{gray}$\pm$ 0.03} & 3.54 {\tiny\color{gray}$\pm$ 0.11} & 3.07 {\tiny\color{gray}$\pm$ 0.10} & 3.5 {\tiny\color{gray}$\pm$ 0.01} \\
           & \texttt{S\&T} & 86.31 {\tiny\color{gray}$\pm$ 0.33} & 86.36 {\tiny\color{gray}$\pm$ 0.29} & 86.81 {\tiny\color{gray}$\pm$ 1.05} & 79.53 {\tiny\color{gray}$\pm$ 1.72} & 85.61 {\tiny\color{gray}$\pm$ 0.59} & 4.69 {\tiny\color{gray}$\pm$ 0.02} & 4.72 {\tiny\color{gray}$\pm$ 0.02} & 4.78 {\tiny\color{gray}$\pm$ 0.05} & 3.82 {\tiny\color{gray}$\pm$ 0.07} & 4.76 {\tiny\color{gray}$\pm$ 0.03} \\
           & \texttt{PivotRR} & 64.42 {\tiny\color{gray}$\pm$ 1.08} & 65.18 {\tiny\color{gray}$\pm$ 0.74} & 61.44 {\tiny\color{gray}$\pm$ 3.10} & 60.56 {\tiny\color{gray}$\pm$ 2.12} & 64.26 {\tiny\color{gray}$\pm$ 1.39} & 3.70 {\tiny\color{gray}$\pm$ 0.04} & 3.68 {\tiny\color{gray}$\pm$ 0.03} & 3.65 {\tiny\color{gray}$\pm$ 0.10} & 3.24 {\tiny\color{gray}$\pm$ 0.07} & 3.76 {\tiny\color{gray}$\pm$ 0.04} \\
           & \texttt{NeutralRR} & 65.82 {\tiny\color{gray}$\pm$ 1.07} & 65.59 {\tiny\color{gray}$\pm$ 0.75} & 63.79 {\tiny\color{gray}$\pm$ 3.25} & 61.90 {\tiny\color{gray}$\pm$ 2.23} & 64.92 {\tiny\color{gray}$\pm$ 1.39} & 3.80 {\tiny\color{gray}$\pm$ 0.03} & 3.78 {\tiny\color{gray}$\pm$ 0.02} & 3.78 {\tiny\color{gray}$\pm$ 0.09} & 3.36 {\tiny\color{gray}$\pm$ 0.06} & 3.84 {\tiny\color{gray}$\pm$ 0.04} \\
\midrule
\multirow{4}{*}{fr} & \texttt{M2MS} & 60.63 {\tiny\color{gray}$\pm$ 2.11} & 63.81 {\tiny\color{gray}$\pm$ 2.92} & 61.62 {\tiny\color{gray}$\pm$ 1.63} & 59.83 {\tiny\color{gray}$\pm$ 5.47} & 59.98 {\tiny\color{gray}$\pm$ 3.05} & 3.57 {\tiny\color{gray}$\pm$ 0.07} & 3.47 {\tiny\color{gray}$\pm$ 0.09} & 3.60 {\tiny\color{gray}$\pm$ 0.05} & 3.04 {\tiny\color{gray}$\pm$ 0.20} & 3.5 {\tiny\color{gray}$\pm$ 0.10} \\
           & \texttt{S\&T} & 86.70 {\tiny\color{gray}$\pm$ 0.38} & 87.27 {\tiny\color{gray}$\pm$ 0.81} & 86.51 {\tiny\color{gray}$\pm$ 0.57} & 81.17 {\tiny\color{gray}$\pm$ 3.12} & 85.42 {\tiny\color{gray}$\pm$ 1.22} & 4.68 {\tiny\color{gray}$\pm$ 0.03} & 4.70 {\tiny\color{gray}$\pm$ 0.05} & 4.76 {\tiny\color{gray}$\pm$ 0.03} & 3.72 {\tiny\color{gray}$\pm$ 0.18} & 4.66 {\tiny\color{gray}$\pm$ 0.06} \\
           & \texttt{PivotRR} & 64.67 {\tiny\color{gray}$\pm$ 2.02} & 66.29 {\tiny\color{gray}$\pm$ 2.93} & 64.62 {\tiny\color{gray}$\pm$ 1.53} & 62.68 {\tiny\color{gray}$\pm$ 4.89} & 62.06 {\tiny\color{gray}$\pm$ 3.14} & 3.74 {\tiny\color{gray}$\pm$ 0.06} & 3.67 {\tiny\color{gray}$\pm$ 0.09} & 3.81 {\tiny\color{gray}$\pm$ 0.05} & 3.27 {\tiny\color{gray}$\pm$ 0.14} & 3.66 {\tiny\color{gray}$\pm$ 0.09} \\
           & \texttt{NeutralRR} & 66.34 {\tiny\color{gray}$\pm$ 1.99} & 66.42 {\tiny\color{gray}$\pm$ 3.29} & 65.32 {\tiny\color{gray}$\pm$ 1.60} & 63.04 {\tiny\color{gray}$\pm$ 5.42} & 64.02 {\tiny\color{gray}$\pm$ 2.98} & 3.86 {\tiny\color{gray}$\pm$ 0.05} & 3.76 {\tiny\color{gray}$\pm$ 0.08} & 3.89 {\tiny\color{gray}$\pm$ 0.05} & 3.25 {\tiny\color{gray}$\pm$ 0.14} & 3.75 {\tiny\color{gray}$\pm$ 0.07} \\
\midrule
\multirow{5}{*}{zh} & \texttt{M2MS} & 61.95 {\tiny\color{gray}$\pm$ 1.16} & 59.45 {\tiny\color{gray}$\pm$ 2.13} & 61.08 {\tiny\color{gray}$\pm$ 5.42} & 60.23 {\tiny\color{gray}$\pm$ 1.05} & 60.76 {\tiny\color{gray}$\pm$ 2.04} & 3.40 {\tiny\color{gray}$\pm$ 0.04} & 3.21 {\tiny\color{gray}$\pm$ 0.07} & 3.41 {\tiny\color{gray}$\pm$ 0.16} & 3.20 {\tiny\color{gray}$\pm$ 0.01} & 3.4 {\tiny\color{gray}$\pm$ 0.10} \\
           & \texttt{S\&T} & 83.61 {\tiny\color{gray}$\pm$ 0.53} & 82.72 {\tiny\color{gray}$\pm$ 1.54} & 83.48 {\tiny\color{gray}$\pm$ 3.32} & 82.50 {\tiny\color{gray}$\pm$ 0.57} & 81.42 {\tiny\color{gray}$\pm$ 1.40} & 4.26 {\tiny\color{gray}$\pm$ 0.04} & 4.22 {\tiny\color{gray}$\pm$ 0.06} & 4.30 {\tiny\color{gray}$\pm$ 0.13} & 4.10 {\tiny\color{gray}$\pm$ 0.02} & 4.31 {\tiny\color{gray}$\pm$ 0.06} \\
           & \texttt{Mistral 7B} & 67.28 {\tiny\color{gray}$\pm$ 1.27} & 68.27 {\tiny\color{gray}$\pm$ 1.66} & 70.39 {\tiny\color{gray}$\pm$ 4.04} & 66.40 {\tiny\color{gray}$\pm$ 0.86} & 65.42 {\tiny\color{gray}$\pm$ 1.67} & 3.19 {\tiny\color{gray}$\pm$ 0.06} & 3.07 {\tiny\color{gray}$\pm$ 0.09} & 3.44 {\tiny\color{gray}$\pm$ 0.23} & 2.98 {\tiny\color{gray}$\pm$ 0.05} & 3.08 {\tiny\color{gray}$\pm$ 0.09} \\
           & \texttt{PivotRR} & 64.73 {\tiny\color{gray}$\pm$ 1.13} & 60.92 {\tiny\color{gray}$\pm$ 2.15} & 61.59 {\tiny\color{gray}$\pm$ 5.66} & 62.99 {\tiny\color{gray}$\pm$ 1.01} & 62.33 {\tiny\color{gray}$\pm$ 2.04} & 3.54 {\tiny\color{gray}$\pm$ 0.04} & 3.37 {\tiny\color{gray}$\pm$ 0.06} & 3.56 {\tiny\color{gray}$\pm$ 0.15} & 3.37 {\tiny\color{gray}$\pm$ 0.03} & 3.60 {\tiny\color{gray}$\pm$ 0.05} \\
           & \texttt{NeutralRR} & 66.94 {\tiny\color{gray}$\pm$ 0.96} & 62.77 {\tiny\color{gray}$\pm$ 2.20} & 62.83 {\tiny\color{gray}$\pm$ 6.14} & 63.74 {\tiny\color{gray}$\pm$ 1.03} & 63.51 {\tiny\color{gray}$\pm$ 1.88} & 3.63 {\tiny\color{gray}$\pm$ 0.03} & 3.47 {\tiny\color{gray}$\pm$ 0.06} & 3.64 {\tiny\color{gray}$\pm$ 0.14} & 3.43 {\tiny\color{gray}$\pm$ 0.03} & 3.66 {\tiny\color{gray}$\pm$ 0.05} \\
\midrule
\multirow{4}{*}{rest} & \texttt{M2MS} & 60.21 {\tiny\color{gray}$\pm$ 0.98} & 62.63 {\tiny\color{gray}$\pm$ 1.33} & 60.04 {\tiny\color{gray}$\pm$ 3.37} & 60.86 {\tiny\color{gray}$\pm$ 1.94} & 60.87 {\tiny\color{gray}$\pm$ 1.22} & 3.56 {\tiny\color{gray}$\pm$ 0.03} & 3.45 {\tiny\color{gray}$\pm$ 0.04} & 3.51 {\tiny\color{gray}$\pm$ 0.10} & 3.22 {\tiny\color{gray}$\pm$ 0.06} & 3.59 {\tiny\color{gray}$\pm$ 0.04} \\
           & \texttt{S\&T} & 85.05 {\tiny\color{gray}$\pm$ 0.26} & 85.82 {\tiny\color{gray}$\pm$ 0.54} & 85.72 {\tiny\color{gray}$\pm$ 1.17} & 80.39 {\tiny\color{gray}$\pm$ 1.35} & 85.22 {\tiny\color{gray}$\pm$ 0.53} & 4.72 {\tiny\color{gray}$\pm$ 0.02} & 4.75 {\tiny\color{gray}$\pm$ 0.03} & 4.72 {\tiny\color{gray}$\pm$ 0.05} & 3.96 {\tiny\color{gray}$\pm$ 0.06} & 4.78 {\tiny\color{gray}$\pm$ 0.03} \\
           & \texttt{PivotRR} & 63.02 {\tiny\color{gray}$\pm$ 0.98} & 64.63 {\tiny\color{gray}$\pm$ 1.34} & 61.47 {\tiny\color{gray}$\pm$ 3.15} & 62.33 {\tiny\color{gray}$\pm$ 2.03} & 63.02 {\tiny\color{gray}$\pm$ 1.20} & 3.73 {\tiny\color{gray}$\pm$ 0.03} & 3.69 {\tiny\color{gray}$\pm$ 0.04} & 3.68 {\tiny\color{gray}$\pm$ 0.09} & 3.37 {\tiny\color{gray}$\pm$ 0.06} & 3.77 {\tiny\color{gray}$\pm$ 0.04} \\
           & \texttt{NeutralRR} & 64.90 {\tiny\color{gray}$\pm$ 0.94} & 66.30 {\tiny\color{gray}$\pm$ 1.27} & 62.98 {\tiny\color{gray}$\pm$ 3.15} & 63.86 {\tiny\color{gray}$\pm$ 1.98} & 64.42 {\tiny\color{gray}$\pm$ 1.21} & 3.83 {\tiny\color{gray}$\pm$ 0.03} & 3.78 {\tiny\color{gray}$\pm$ 0.04} & 3.78 {\tiny\color{gray}$\pm$ 0.08} & 3.45 {\tiny\color{gray}$\pm$ 0.05} & 3.86 {\tiny\color{gray}$\pm$ 0.04} \\
\bottomrule
\end{tabular}
\caption{Extended results of multi-target cross-lingual summarization in CrossSum for the metrics evaluating semantic coherence across target languages.}
\label{tab:extended_results_coherence}
\end{table}
\end{landscape}

\end{document}